\documentclass[]{fairmeta}

\usepackage{microtype}
\usepackage{graphicx}
\usepackage{booktabs} 
\usepackage{adjustbox}

\usepackage{hyperref}
\usepackage{arydshln}
\usepackage{makecell}

\usepackage{amsmath}
\usepackage{amssymb}
\usepackage{mathtools}
\usepackage{amsthm}
\usepackage{multicol}
\usepackage{multirow}
\usepackage{resizegather}

\usepackage{natbib}
\usepackage{enumitem}
\usepackage{subcaption}
\usepackage{wrapfig}
\usepackage{array}
\usepackage{algorithm}
\usepackage{algpseudocode}
\usepackage{xcolor}
\usepackage{colortbl}
\newcommand{\bench}[2]{\mbox{#1~\citep{#2}}}

\theoremstyle{plain}

\theoremstyle{definition}

\theoremstyle{remark}

\usepackage[textsize=tiny]{todonotes}

\title{One Step, Four Clocks: How GUI Agents Fall Out of Step}

\author[1]{Zhanwen Chen}
\author[2]{Tianchun Wang$^{\#}$}

\affiliation[1]{Hadron}
\affiliation[2]{Johns Hopkins University}

\abstract{
GUI agents can choose correct actions yet fail as interfaces change during deliberation. Latency alone cannot explain these failures because actions may become invalid at different interaction stages. We model four events: environment change, observation availability, decision commitment, and action receipt (EODA). We introduce \textbf{ChronotopeGym} to study these failures under Paused, Replay, and Live execution. Across five tasks, agents can fail in Live despite succeeding when responses remain valid, even when actions meet deadlines. On Moving Balloon, updating a pending click raises success from 0\% to 100\% without changing response time. Our accelerator-resident GUI controller, \textbf{Chronotope}, maintains compact interaction history and selectively keeps, updates, or redecides pending actions. Compared with Agent~S3, Chronotope reduces median end-to-end latency by 85.5\% for a 27B model, expanding its performance envelope to 100\% Live success at an 8-second deadline versus 0\%. Chronotope enables real-time GUI control through compact interaction history, knowing when to preserve, revise, or replace actions, and an overall efficient architectural design that obviates verbose reflection, reasoning, and grounding calls.
}

\metadata[$^{\#}$Correspondence]{Tianchun Wang at \email{tcwangai@gmail.com}}
\code{\href{https://github.com/hadronresearch/chronotope}{https://github.com/hadronresearch/chronotope}}

\begin{document}
\maketitle

\begin{figure}[ht]
    \centering
    \includegraphics[width=\linewidth]{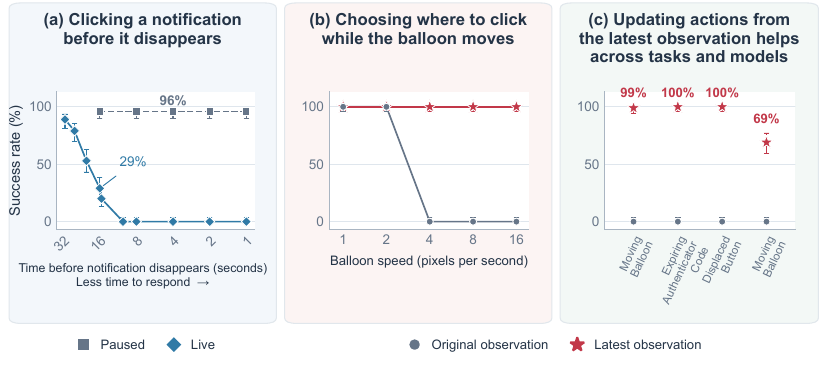}
    \caption{When a screen keeps changing, agents must act in time and update their actions to match it. \textbf{(a)} In Disappearing Notification, an Agent S3 model must click a notification before it disappears. We shorten this time limit and compare success when the interface waits while the agent thinks (Paused) against when it keeps running (Live). Success stays at 96\% in Paused but falls in Live as notifications disappear sooner, showing how pausing can hide timing failures. \textbf{(b)} In Moving Balloon, a Qwen3.8-27B agent must click a moving balloon. We increase its speed and compare clicks aimed at its original versus latest observed positions. The model call takes the same time in both cases. At higher speeds (4, 8, and 16 pixels/s), updating the position raises success from 0\% to 100\%. \textbf{(c)} We test the same update across three tasks. With Qwen3.8-27B, success rises from 0\% to 99--100\%, showing that the update helps with both changing positions and changing text. The update method generalizes to another model, UI-TARS-1.5-7B, whose success increased from 0\% to 69\%. Each point represents 100 trials, and error bars 95\% confidence intervals. See \S\ref{sec:experiments} for experiment details.
}\label{fig:headline_results}
\end{figure}

\section{Introduction}\label{intro}

\begin{figure}[ht]
\centering
\begin{subfigure}[t]{0.30\textwidth}
    \centering
    \includegraphics[width=\linewidth]{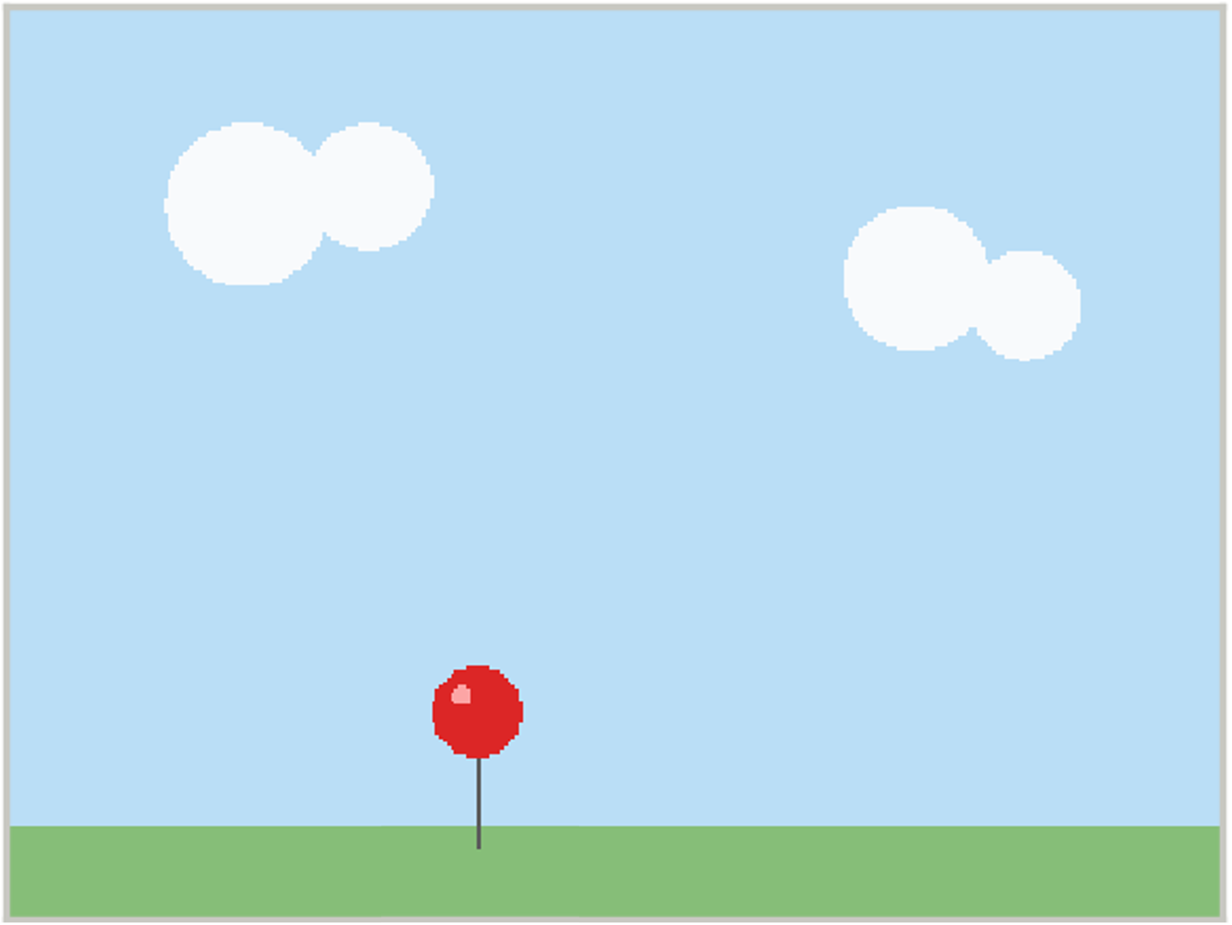}
    \caption{The agent sees the balloon at its original position}
    \label{fig:hero_observe}
\end{subfigure}
\hfill
\begin{subfigure}[t]{0.30\textwidth}
    \centering
    \includegraphics[width=\linewidth]{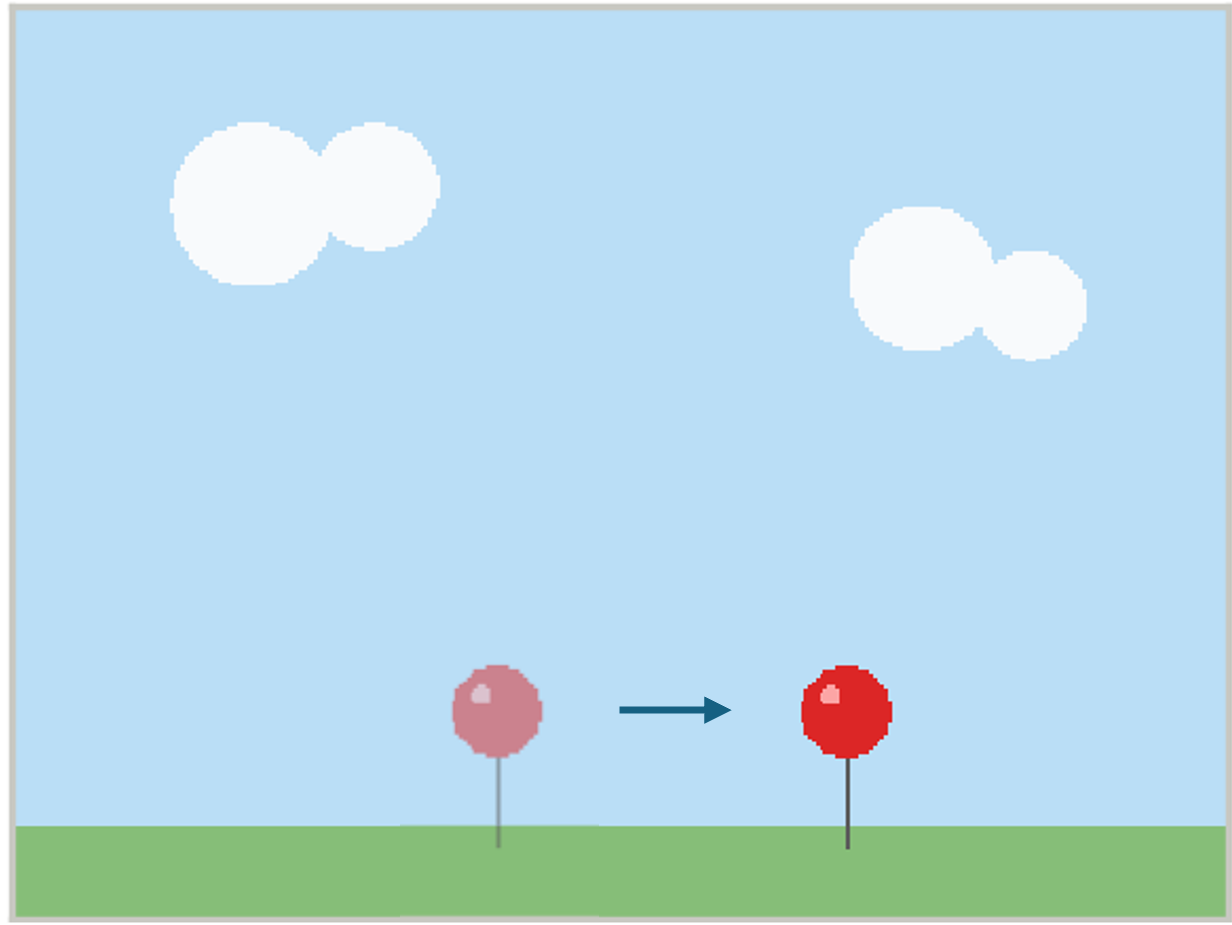}
    \caption{The balloon moves while the agent reasons}
    \label{fig:hero_deliberate}
\end{subfigure}
\hfill
\begin{subfigure}[t]{0.30\textwidth}
    \centering
    \includegraphics[width=\linewidth]{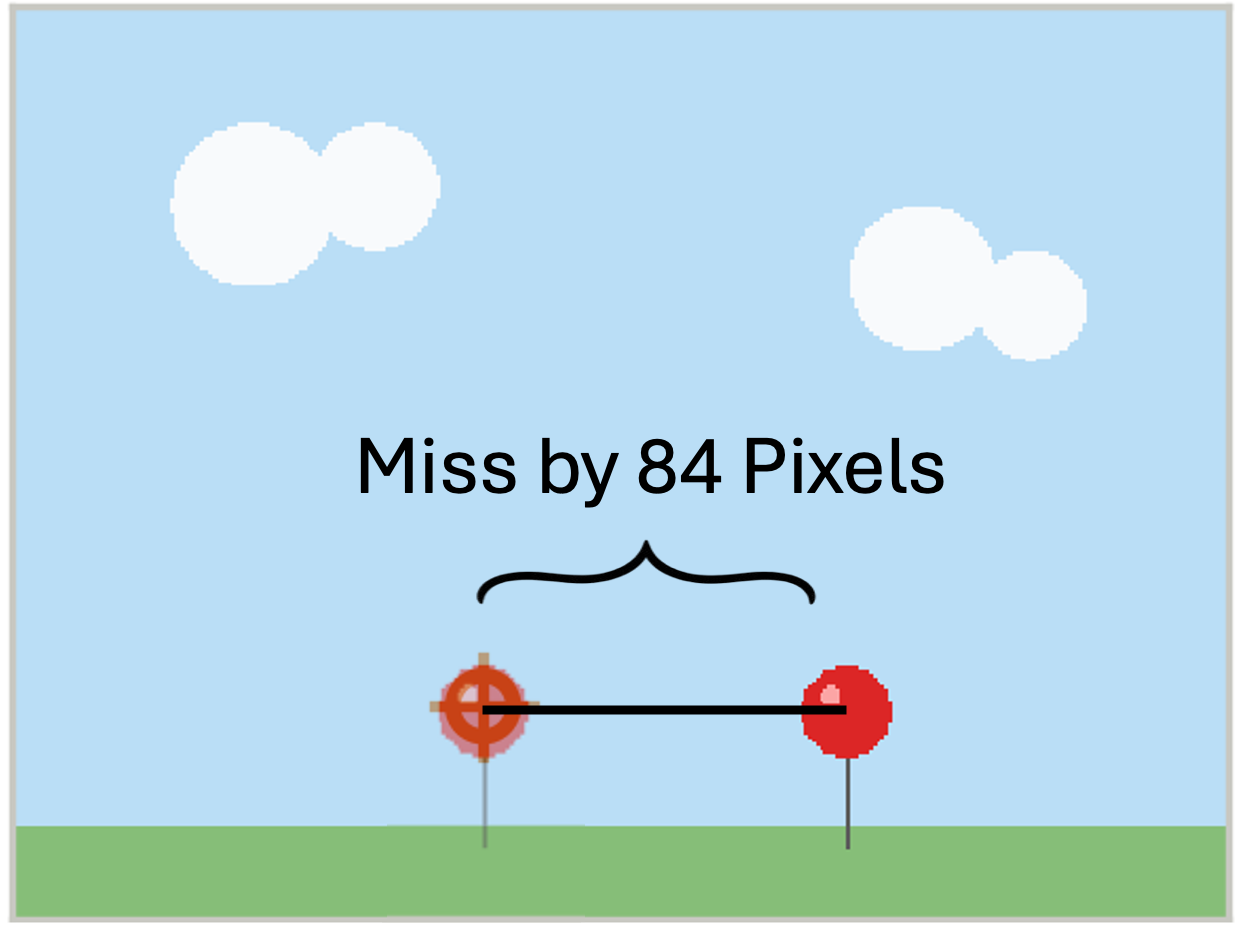}
    \caption{The agent's click misses the balloon by 84 pixels}
    \label{fig:hero_act}
\end{subfigure}
\caption{\textbf{Correct for the observed screen, wrong on arrival.} In Moving Balloon, the agent must click a moving balloon. The agent chooses the correct position for its initial observation, but the interface continues evolving while it computes, so the same position is wrong when the click arrives. The faded balloon marks the initially observed position. The crosshair marks where the click lands.}
\label{fig:hero}
\end{figure}

Graphical user interface (GUI) agents are typically evaluated as if the interface waits for them. The agent observes a screen, thinks, produces an action, and is judged on whether that action was correct \citep{deng2023mind2web, zhou2024webarena, zheng2024seeact, rawles2025androidworld,xie2024osworld, qin2025uitars, li2026aoi}. Real interfaces continue evolving independently. A page may finish loading, a notification may disappear, a target may move away from its observed position, or an earlier click may keep the player character advancing while the agent reacts to enemy fire. A click can thus be right for the frame the agent saw and wrong for the state in which it lands. Figure~\ref{fig:hero} illustrates this with a moving balloon: the agent chooses the correct click position for its initial image, but the click misses because the balloon has moved by the time it arrives. This reveals a gap between solving a GUI task when the interface waits and solving it while the interface continues to evolve.

\begin{figure}[t]
    \centering
    \includegraphics[width=\textwidth]{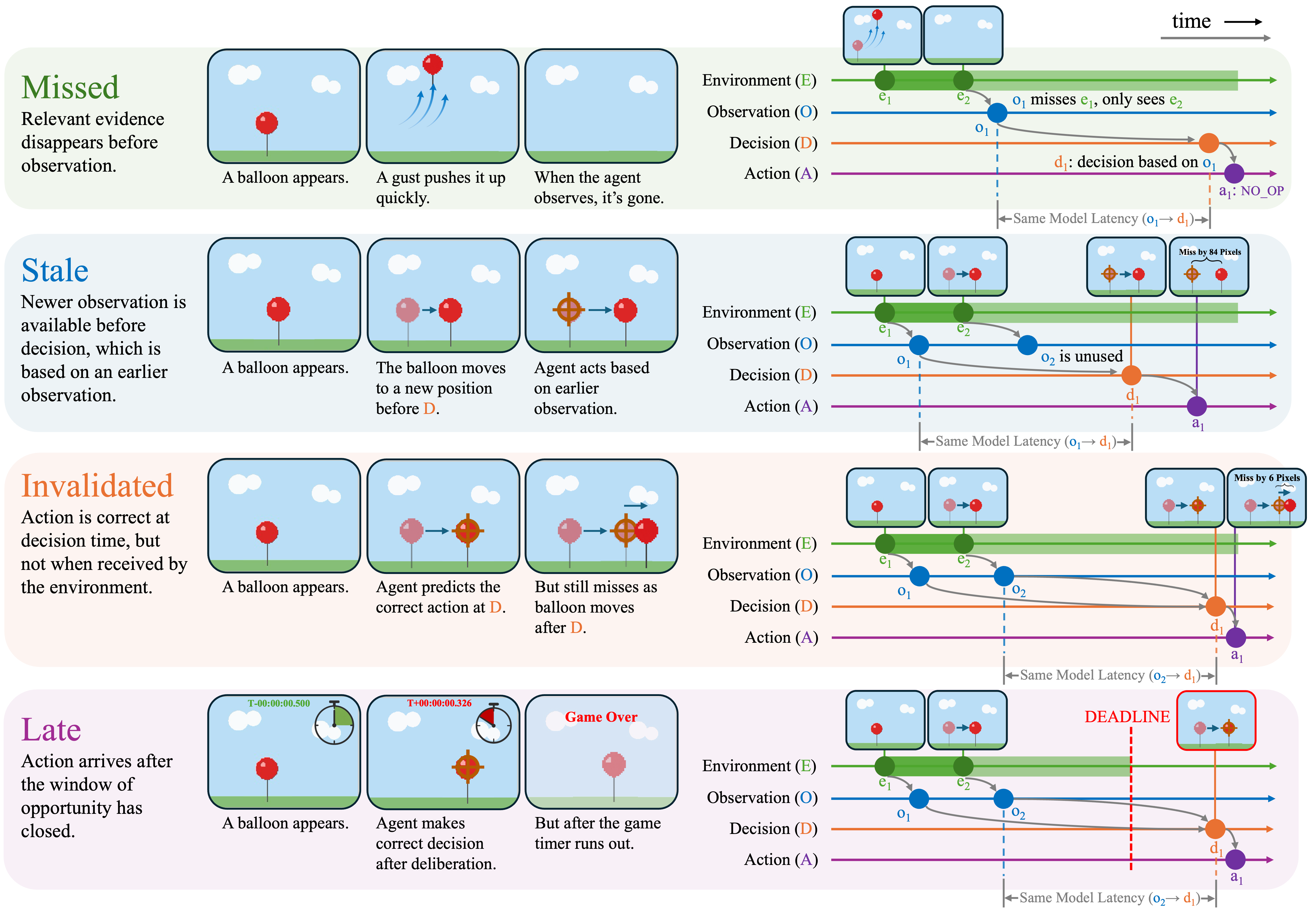}
    \caption{\textbf{The same model latency conceals four timing failures.} Four Moving Balloon examples share the same observation-to-decision interval but fail for different reasons. \textbf{Missed.} Relevant evidence disappears before the agent observes it. \textbf{Stale.} Newer evidence is available before commitment. The agent still uses an earlier observation. \textbf{Invalidated.} The action is correct when committed. It becomes wrong before the application receives it. \textbf{Late.} The action reaches the application after the response deadline. The timelines mark environment changes ($E$), observation availability ($O$), decision commitment ($D$), and application receipt ($A$). They show which observation supports each decision. \textbf{\textit{Model latency alone does not decide timing failure.}}}\label{fig:eoda_latency_not_enough_4failures}
\end{figure}

A natural explanation is that the agent is too slow. Sometimes the action simply reaches the application after the deadline. But model latency alone does not tell us what went wrong, because a response does not fail at a single moment. It unfolds through a lifecycle in which the interface changes, information becomes available to the controller, a decision becomes fixed, and the resulting action reaches the application. Correctness can be lost at different points along this lifecycle, even when the model takes the same amount of time. Figure~\ref{fig:eoda_latency_not_enough_4failures} shows that the the same model latency can cause different failures. Diagnosing a timing failure therefore requires knowing not only how long computation took, but where the response correctness is lost.

We describe this lifecycle using four events. \(E\) marks when a task-relevant environment change occurs, \(O\), when an observation becomes available to the controller, \(D\), when the action can no longer be changed, and \(A\), when the application receives the action. Figure~\ref{fig:eoda_latency_not_enough_4failures} illustrates four timing failure modes: \begin{itemize}[nosep, topsep=-0.45\baselineskip, after=\vspace{-0.05\baselineskip}]
    \item \textbf{Missed}. Relevant evidence disappears before it is observed.
    \item \textbf{Stale}. The agent acts based on an outdated observation when newer evidence is available.
    \item \textbf{Invalidated}. A correct action at \(D\) becomes wrong before the application receives it at \(A\).
    \item \textbf{Late}. A response reaches \(A\) after its opportunity has closed.
\end{itemize} A single latency number tells how long a response took. It cannot tell us where correctness was lost.

The four failures require different remedies. A missed event requires observation to continue while the agent deliberates. But continued observation is not enough to prevent a stale decision if new evidence can affect only the next decision. Relevant evidence arriving before \(D\) must therefore be able to change the decision already underway. Once an action is committed, an unexpected change after \(D\) cannot be repaired by revising that decision, and a late response still has to reach \(A\) sooner. We investigate these failures and their remedies in three steps: \begin{itemize}[nosep, topsep=-0.45\baselineskip, after=\vspace{-0.01\baselineskip}]
    \item \textbf{Control the temporal structure of GUI interaction}. ChronotopeGym evaluates tasks under three conditions: Paused, Replay, and Live. Paused freezes task progression during deliberation. Replay and Live deliver the same observations on the same schedule, but only Live allows the conditions for a successful response to change during deliberation.
    \item \textbf{Isolate the effect of newer evidence}. We change the observation supporting a pending decision while holding decision start, commitment, and application receipt fixed. This separates the benefit of newer evidence from that of a faster response.
    \item \textbf{Check pending decisions before commitment}. Chronotope continues observing during deliberation, maintains the current control state and pending decision, and checks that decision against the newest relevant evidence before commitment.
\end{itemize}

Our diagnosis and control approach are both based on the EODA framework. Specifically, \textbf{ChronotopeGym} is a test bed for diagnosing and evaluating timing failures in GUI agents. It uses EODA to separate the timing of environment change, observation, decision, and action receipt during GUI interaction. It changes when evidence becomes available, when a decision becomes fixed, and when an action reaches the application, allowing us to test where correctness is lost. \textbf{Our approach, Chronotope,} keeps an unfinished decision current as the interface changes, preserving what remains correct and updating information that has become outdated before commitment. \textbf{Across five temporally dynamic GUI tasks, our experimental results show that these timing distinctions matter.} A planner-grounder agent drops from \textbf{\textit{96\% success in Replay to 29\% in Live}} when the same evidence arrives at the same times but the response opportunity continues advancing. On the Moving Balloon task, using an updated observation to locate the target \textbf{\textit{raises success from 0\% to 100\%}}. Our contributions are as follows:




\begin{itemize}
    \item We introduce the EODA temporal framework and the ChronotopeGym benchmark to identify and systematically investigate temporal failures in GUI agents, demonstrating that correct and timely actions can fail when environments change during deliberation.

    \item We propose Chronotope, an easy-to-use GUI agent controller that expands the performance envelope of traditional GUI agents, enabling the same VLM models to achieve 100\% success rate on latency-sensitive tasks where they otherwise result in 0\% under tight deadlines. 

    \item Chronotope's efficient control architecture substantially reduces redundant model-side computation through compact interaction history representation, decision meta-control, and GPU-resident execution. Its overall efficiency leads to 85.5\% lower end-to-end latency and up to 87.5\% lower token consumption than Agent S3.
\end{itemize}


\section{Related Work}\label{related_works}

GUI agents solve computer tasks by observing the screen and taking actions~\citep{qin2025uitars, gonzalez2026scaling}, but an interface can change before an action takes effect \mbox{\citep{liu2026dynamicgui, yao2026livingscreen}}. We review how existing benchmarks evaluate this timing problem and how latency control methods address it.

\textbf{Time in GUI-agent benchmarks}. Recent works study timing in GUI interaction with benchmark-specific tasks and metrics~\citep{liu2026dynamicgui, yao2026livingscreen, li2026aoi, dong2026aapt}. However, a benchmark score alone does not tell us whether the agent failed due to missed changes, outdated information, or late actions. We address this gap with an event-stream view that diagnoses temporal failures across environment change, observation, decision commitment, and action receipt, enabling controlled studies under a common measurement scheme. \S\ref{appendix:review_bench_time} compares eighteen benchmarks on these criteria.

\textbf{Latency control in GUI agents}. Latency can cause GUI agents to act on outdated information, miss action windows, or respond too late to unexpected changes. Existing methods supply newer evidence through AOI~\citep{li2026aoi}, predict future states through MobileDreamer and TraceR1~\citep{cao2026mobiledreamer, liang2026anticipatory}, or prepare and reuse computation before execution through UFO2 and AAPT~\citep{zhang2025ufo2, dong2026aapt}. However, newer evidence may go unused, and planning ahead alone does not update where the agent clicks or types when the screen changes. As a result, an agent may preserve the right high-level intent but still execute an outdated or mistimed action. Our event-stream view identifies where these failures arise and which interventions address them. This enables targeted control at the stage where timing causes correctness to break down. \S\ref{appendix:latency_control} and Table~\ref{tab:review} provide extended discussion and comparisons.

\section{Where is Correctness Lost?}\label{sec:chronotope-gym}

Existing methods cannot identify where temporal correctness is lost between observation and action. We dissect this process into environment change \(E\), observation availability \(O\), decision commitment \(D\), and action receipt \(A\), allowing us to localize where an otherwise valid response becomes incorrect. To achieve this goal, we build \textbf{ChronotopeGym} as a diagnostic test bed that controls the timing of these events.

\subsection{From Observation-Action to EODA}\label{sec:eoda}

\begin{figure}[t]
    \centering
    \includegraphics[width=\linewidth]{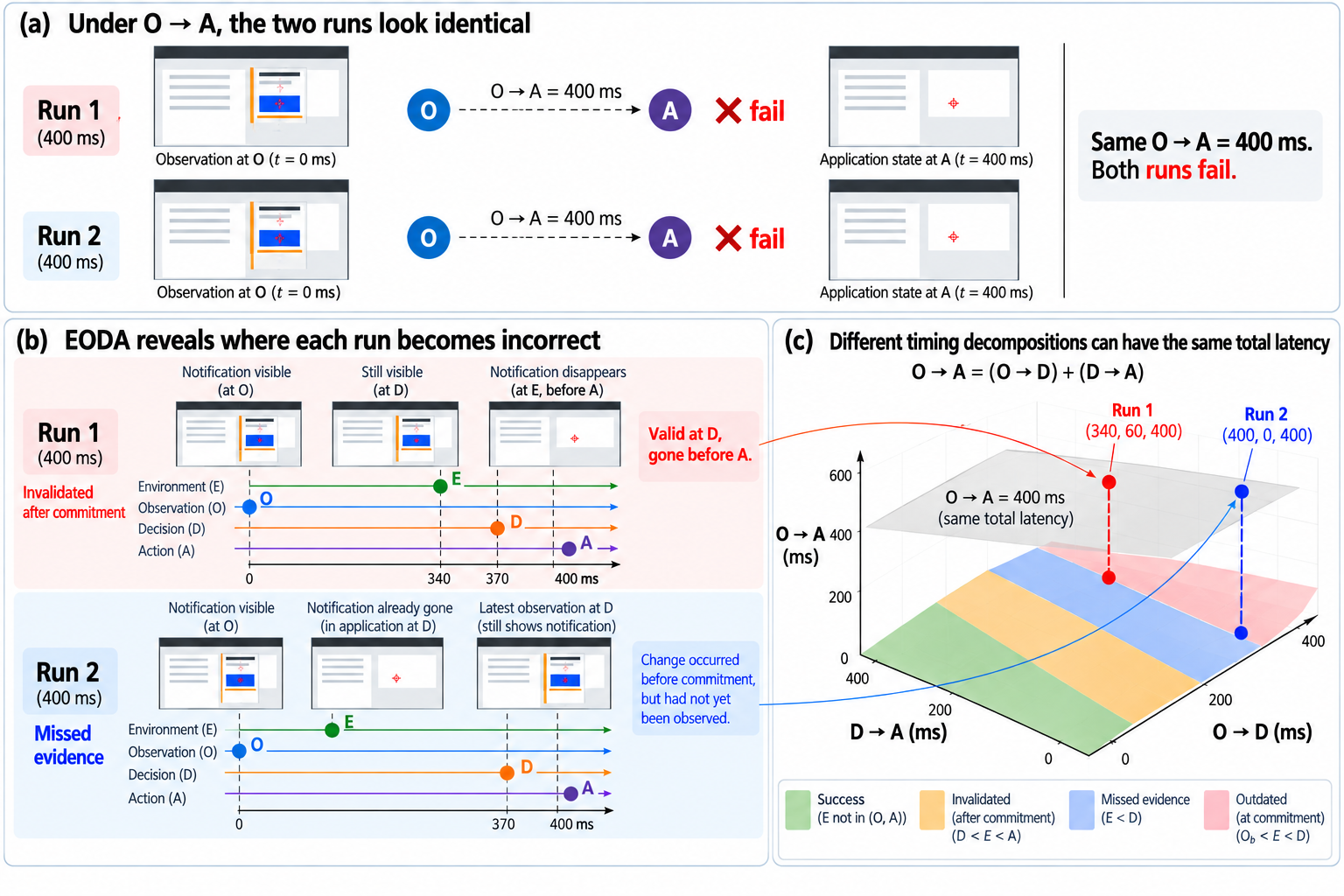}
    \caption{
    \textbf{EODA distinguishes failures hidden by the same observation-to-action latency.}
    Data are from two runs of the \textbf{Disappearing Notification} experiment in ChronotopeGym.
    \textbf{(a)} Run~1 and Run~2 both have \(O\!\rightarrow\!A=400\)~ms latency and both fail, so observation-to-action timing treats them identically.
    \textbf{(b)} ChronotopeGym records the environment-change event \(E\) when the notification disappears, allowing EODA to localize where each failure occurs.
    In Run~1, \(E\) occurs after decision commitment \(D\) but before action receipt \(A\), so the committed action is invalidated before execution.
    In Run~2, \(E\) occurs before \(D\), but the latest observation available at commitment still shows the notification, so the agent misses relevant evidence.
    These diagnoses indicate different mitigation targets: reducing action-delivery delay in the first case and obtaining or incorporating fresher observations in the second.
    \textbf{(c)} Generally, even when \(O\!\rightarrow\!A\) is fixed at 400~ms, different allocations of time to computation \(O\!\rightarrow\!D\) and delivery \(D\!\rightarrow\!A\) can produce different failure modes.
    }
    \label{fig:oa-eoda-timing}
\end{figure}

To identify why the same response time can produce different failures, we separate the events hidden within observation-to-action latency. Figure~\ref{fig:oa-eoda-timing} shows two \textbf{Disappearing Notification} executions with the same \(O\!\rightarrow\!A=400\) ms latency. In the first, the decision is committed while the notification is still present, but the notification disappears before the click arrives. In the second, the notification disappears before commitment, but that change has not yet appeared in the agent's latest observation. Observation-to-action timing assigns the same latency to both executions, even though correctness is lost at different stages. By collapsing these distinct failure mechanisms into a single end-to-end measurement, it cannot identify which part of the response caused the failure.

We therefore dissect the response into four events: environment change \(E\), observation availability \(O\), decision commitment \(D\), and action receipt \(A\). This decomposition exposes when relevant state changes relative to what the agent observes, decides, and executes. As Figure~\ref{fig:oa-eoda-timing}(c) shows, the same \(O\!\rightarrow\!A\) latency can contain different \(O\!\rightarrow\!D\) and \(D\!\rightarrow\!A\) intervals. EODA thus distinguishes whether correctness is lost because relevant evidence was unavailable to the decision, because the decision became outdated, or because the environment changed after commitment but before action receipt.

\subsection{Controlling Timing with Paused, Replay, and Live}\label{sec:temporal-conditions}

In order to control timing within EODA, we introduce three execution conditions, \textbf{Paused}, \textbf{Replay}, and \textbf{Live}. In Paused, task-relevant state is held fixed while the agent computes and acts. In Replay, the agent receives the same changing observations as in a corresponding Live execution at the same cadence, but the response opportunity remains valid until the agent acts. In Live, both the observations and the underlying task state continue advancing while the agent computes and acts. These conditions let us vary whether task evolution is observed and whether that evolution can invalidate the response.

Comparing the three conditions separates different effects of task evolution. Paused versus Replay tests whether changing observations alone affect the agent when the response remains valid. Replay versus Live then tests what changes when the same evolving evidence is coupled to an underlying task that can invalidate the response before action receipt. Paused, Replay, and Live therefore let us distinguish failures caused by processing evolving evidence from failures caused by the task continuing to change while the agent responds.

We choose five tasks to represent different ways a GUI response can become invalid over time: \textbf{Disappearing Notification}, where the opportunity to respond closes, \textbf{Moving Balloon}, where the target location changes, \textbf{Expiring Authenticator Code}, where an observed value becomes invalid, \textbf{Displaced Button}, where the location of the intended control changes, and \textbf{Reversing Trading Signal}, where the required action itself changes. Together, they cover changes to the opportunity, target, value, control location, and action rather than particular application domains. Appendix~\ref{appendix:task-protocol} details task definitions, action spaces, success criteria, and implementation. After Paused, Replay, and Live establish whether task evolution changes the outcome, we use finer EODA interventions to vary which observation can affect the decision, when the decision is committed, or when the action is received while holding the remaining timing fixed. This lets us localize whether correctness is lost from missed evidence, an outdated decision at commitment, or invalidation after commitment.



\section{Chronotope: Towards Control in Physical Time}\label{sec:chronotope}

\begin{figure}[htb]
    \centering
    \includegraphics[width=\textwidth]{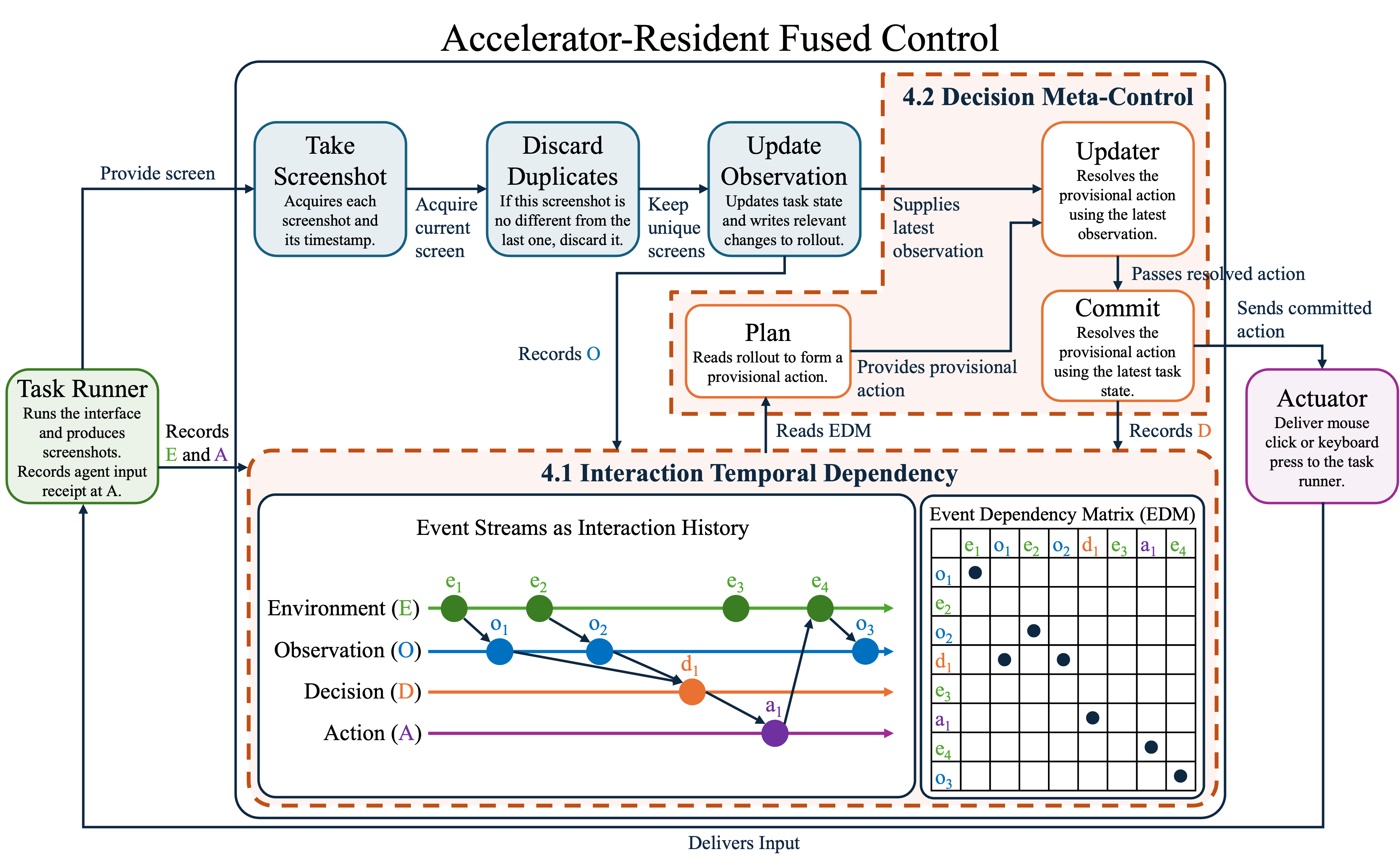}
    \caption{\textbf{Chronotope's accelerator-resident fused control pipeline} (\S\ref{sec:fused-execution}). The controller discards duplicate screenshots, updates the interaction history, and supplies the latest observation directly to the updater. The planner reads this history to propose an action, which the updater reconciles with the latest observation. The Commit stage finalizes the action at \(D\). The actuator delivers the committed input to the task runner, which records environment changes (\(E\)) and action receipts (\(A\)) in the interaction history. The interaction history comprises \(EODA\) event streams and an Event Dependency Matrix (EDM), where each cell indicates that the row event directly depends on the column event.}\label{fig:fused-control}
\end{figure}

\S\ref{sec:chronotope-gym} uses EODA to identify where GUI responses become incorrect as the interface changes. Its diagnosis reveals that sequential GUI control cannot incorporate observations arriving during inference into a decision already underway. To address this limitation, we introduce \textbf{Chronotope}, a clock-decoupled controller that plans from EODA interaction history while continuously monitoring for new observations, updates provisional actions with newer observations before commitment, and keeps this control path on the accelerator (Figure~\ref{fig:fused-control}). Its interaction history relates new observations to unfinished decisions and previously committed actions (\S\ref{sec:interaction-state}), enabling decision meta-control to determine whether to keep a proposed action, update its arguments, or restart inference before commitment (\S\ref{sec:precommit-meta-control}). Appendix~\ref{appendix:chronotope-algorithm} specifies the Chronotope algorithm and analyzes its time complexity, space complexity, and communication overhead.

\subsection{Interaction Temporal Dependency}\label{sec:interaction-state}

GUI agents interact with a changing environment by observing screenshots and issuing actions. Each screenshot captures the interface state at the time of capture but not the history of observations, decisions, and actions preceding it. An agent using screenshots without linking them to its decisions and actions would fail to capture temporal dependencies during inference. For example, the agent may still be choosing its next action based on a screenshot when an earlier click reaches the application. The click or an unrelated event may change the interface and leave the unfinished decision using an outdated target position or input value. To plan new actions from prior interactions and update decisions as new observations arrive, Chronotope maintains an interaction history that relates these observations to unfinished decisions and previously committed actions.

To represent both when events occur and which events depend on others, Chronotope organizes this interaction history as EODA event streams and their Event Dependency Matrix (EDM) representation shown in \ref{sec:interaction-state} of Figure~\ref{fig:fused-control}. The EDM records direct dependencies between events. For example, decision \(d_1\) uses both observations \(o_1\) and \(o_2\). As new observations arrive, the Update Observation stage records observation events (\(O\)) and relevant state changes in the interaction history. The history also links an action receipt to the resulting environment change and its observation, as in \(a_1 \rightarrow e_4 \rightarrow o_3\). The planner reads this history to propose an action, which the updater checks against the latest relevant observation before commitment.

\subsection{Decisions Meta Control}\label{sec:precommit-meta-control}

An interface change may require a new click coordinate or input value while leaving the intended operation unchanged. For example, in Moving Balloon, the agent should still click the same balloon, but its movement can make the proposed click coordinate outdated. Another example is Expiring Authenticator Code, where the agent should still type into the same field, but a code change can make the proposed value outdated. One naïve solution is \textit{restarting inference} with the new observation, where the controller asks the planner model again what to do. However, only the click coordinate or input value needs updating, wasting inference tokens and incurring the latency of extra model calls. Chronotope instead preserves the intended operation and updates only its outdated arguments before commitment, conserving tokens and reducing latency.

To allow argument updates without restarting inference, Chronotope keeps proposed actions open to revision until commitment (shown as \ref{sec:precommit-meta-control} inside Figure~\ref{fig:fused-control}). The planner proposes an action, and the updater checks its intended operation and arguments against the latest relevant observation. If both remain valid, Chronotope keeps the action unchanged. If only an argument is outdated, revision replaces it while preserving the operation. Instead of specifying an argument early and revising it later, late binding leaves a state-dependent argument unspecified and resolves it from the newest relevant observation just before commitment. If the operation is no longer appropriate, updating its arguments is insufficient, so re-decision restarts inference. Once the operation and arguments are resolved, the Commit stage finalizes the the action and records it in the interaction history.

\subsection{Accelerator-Resident Fused Control}\label{sec:fused-execution}

\begin{figure}
    \includegraphics[width=\linewidth]{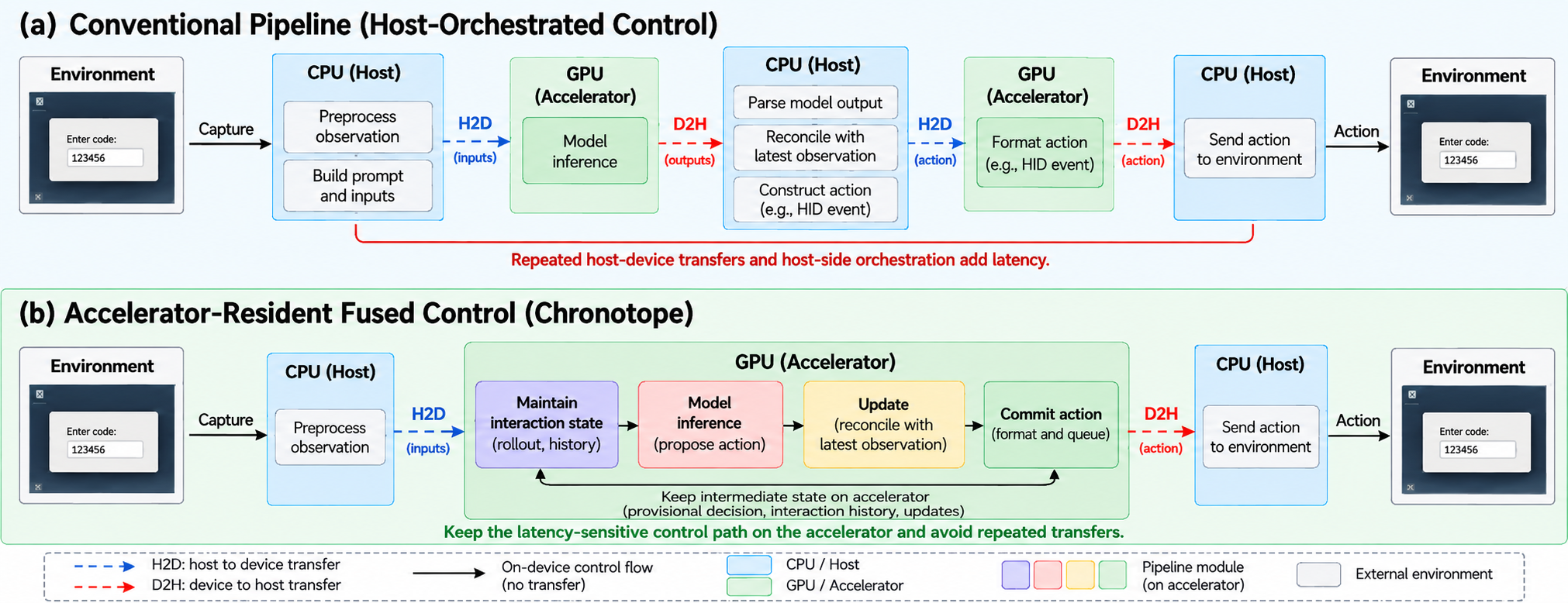}
    \caption{\textbf{Host-orchestrated (a) versus accelerator-resident (b) control.} Chronotope keeps the decision, interaction history, and meta-control operations on the accelerator, reducing host-device transfers represented by dashed arrows.}
    \label{fig:host-resident-control}
\end{figure}
    

Updating an unfinished decision is useful only if the resulting action can still arrive in time. Maintaining interaction state and reconciling a decision introduce additional computation while the interface continues to change. A correct update can therefore still fail if this added work delays the action beyond its useful response window. Chronotope therefore executes history maintenance and decision meta-control on the accelerator to reduce host-side orchestration overhead.

In the host-orchestrated pipeline, model output returns from the GPU to the CPU for parsing, reconciliation, and action construction (Figure~\ref{fig:host-resident-control}(a)). Chronotope instead keeps the provisional decision and interaction history on the accelerator, where it updates the action before committing it at \(D\) (Figure~\ref{fig:host-resident-control}(b)). This reduces control-state transfers and host--device synchronization before commitment. Reducing this overhead can help the updated action reach the application before further interface changes invalidate it.



\section{Experiments}\label{sec:experiments}

To study how interface changes affect GUI-agent success and how agents can respond to these changes, we ask three questions. First, what determines whether an agent can still solve a task when the interface no longer waits for its response? Second, if the interface changes while the agent is still reasoning from earlier observations, how should the agent respond? Third, can an agent respond to interface changes and still deliver its action before the response window closes? We describe the common experimental setup in \S\ref{subsec:experimental-setup} and answer these questions in \S\ref{subsec:claim-1}--\ref{subsec:claim-3}.

\subsection{Experimental Setup}\label{subsec:experimental-setup}

\textbf{Tasks and agents.} We evaluate Chronotope on five temporally dynamic GUI tasks introduced in \S\ref{sec:chronotope-gym}: Disappearing Notification, Moving Balloon, Expiring Authenticator Code, Displaced Button, and Reversing Trading Signal. On the agent controller level, we compare Accelerator-Resident Fused Control (\S\ref{sec:fused-execution}), i.e., Chronotope, against Agent S3~\citep{gonzalez2026scaling}. On the model level, Chronotope primarily uses Qwen3.8-27B~\citep{qwen38}, Gemma-4-31B~\citep{gemmateam2026gemma4}, and UI-TARS-1.5-7B~\citep{qin2025uitars}, while Agent S3~\citep{gonzalez2026scaling} primarily uses EvoCUA-8B~\citep{xue2026evocua} and UI-TARS-1.5-7B~\citep{qin2025uitars}. To separate temporal failure from ordinary task incompetence, each agent configuration enters a timing study only after passing the corresponding Paused qualification check.

\textbf{Study Design and Metrics.} We evaluate each task under the three timing conditions introduced in \S\ref{sec:chronotope-gym}. In \textbf{Paused}, the interface waits while the agent decides. In \textbf{Replay}, the agent receives the same sequence of observations as in Live, at the same times, but the response remains valid while it thinks. In \textbf{Live}, both the observations and the task continue advancing: a target may move, a value may change, or an opportunity may close before the action arrives. We also use \textbf{controlled comparisons} that rerun the same task instance while changing one part of the interaction. We vary which observation supplies the final action or how long a committed action takes to reach the application, while holding the task, model call, and other event times fixed. Each experiment is evaluated on 100 random seeds separate from seeds used in development. We report success and task return at action receipt, together with the \(E\!\rightarrow O\), \(O\!\rightarrow D\), \(D\!\rightarrow A\), and \(E\!\rightarrow A\) intervals.

\textbf{Environment.} All experiments run on one NVIDIA GB10 GPU with local models deployed via vLLM and TensorRT.

\subsection{Claim 1. Live Execution Exposes Failures That Response Time Alone Cannot Explain}\label{subsec:claim-1}

An agent may succeed when the interface waits, but can it do so when the interface keeps changing during its response? If it fails, was the action wrong, or did the changing environment invalidate it before execution? To distinguish these cases, we evaluate the same agent and task under three conditions: Paused, Replay, and Live. In Paused, the interface remains unchanged until the agent responds. In Replay, it evolves as in Live but accepts actions valid for the initial observation. In Live, an action succeeds only if it remains valid upon application receipt. On Disappearing Notification, a baseline Qwen3.8-27B-BF16 agent succeeds in every Paused and Replay trial but fails in every Live trial because its 5.3~s response arrives after the notification's 4~s deadline, despite identifying the correct target. \textit{\textbf{Even a correct action fails if it arrives too late.}}

More surprisingly, an action can fail despite arriving on time. On Moving Balloon and Reversing Trading Signal, success drops from 100\% in Paused and Replay to 0\% in Live despite the agent meeting the response deadline. In Moving Balloon, the agent clicks where the balloon used to be as it moves at 8~pixels/s, so the coordinates are outdated even though clicking remains appropriate. In Reversing Trading Signal, the signal reverses after 2~s, invalidating the agent's original decision.

We isolate the effect of observation freshness from response time by fixing the model call, decision commitment time, and action receipt time, varying only which observation supplies click coordinates. On Moving Balloon, higher speeds (4--16~pixels/s), clicks based on the initial observation always miss, while those based on the latest observation before commitment always hit (Figure~\ref{fig:headline_results}(b)). \textbf{\textit{Even at identical response times, observation freshness can determine whether actions succeed.}}

\subsection{Claim 2. Restarting Inference Can Turn Success Into Failure}\label{subsec:claim-2}

Claim 1 establishes that outdated observations can cause actions to fail, but leaves open how a controller should respond when an observation becomes outdated during inference. A natural response is to restart inference using a newer observation. However, another model call takes time, during which the interface may change again. We compare three responses: keeping the original action, restarting inference, and updating the pending action before commitment (Table~\ref{tab:revision-main}).

Restarting inference can turn a successful action into a failure. On Moving Balloon at 2~pixels/s, Qwen3.8-27B succeeds in every trial by keeping its original action, but restarting inference introduces enough delay to cause every trial to fail. When the balloon's speed doubles to 4~pixels/s, the original click also becomes outdated, causing keeping to fail alongside restarting. Updating the pending click avoids both problems by refreshing its coordinates without restarting inference, succeeding in every trial at both speeds. Restarting only succeeds reliably when the changed state persists, as when the balloon jumps 96~pixels and remains stationary, yielding 99\% success rate. \textbf{\textit{Restarting only works when the interface becomes static, suggesting that agents should update actions instead when the interface keeps changing.}}

Updating an action also requires computation, but its execution cost is small compared with another model call. Chronotope adds $4 \times 10^{-5}$~s of overhead when no update is needed and takes a median of $1.2 \times 10^{-3}$~s to update an action. In contrast, restarting inference with Qwen3.8-27B more than doubles the latency on both Moving Balloon and Expiring Authenticator Code, adding up to 6.1~s. This additional delay can cause the replacement action to become outdated as the interface continues changing, explaining why restarting fails under continuous movement. However, updating action arguments alone cannot fix an outdated decision. On Reversing Trading Signal, the signal itself reverses, requiring the controller to reconsider the intended operation. \textbf{\textit{Updating pending actions avoids the cost and staleness risk of restarting inference, but changes to the intended operation still require a new decision.}}



\begin{table}[ht]
\centering
\small
\setlength{\tabcolsep}{3pt}
\renewcommand{\arraystretch}{1.15}
\begin{tabular}{@{}llccc@{}}
\toprule
\textbf{Task} & \textbf{Setting} & \makecell{\textbf{Keep}\\\textbf{action}} & \makecell{\textbf{Restart}\\\textbf{model}} & \makecell{\textbf{Update}\\\textbf{action}} \\
\midrule
\multicolumn{5}{@{}l}{\textit{Qwen3.8-27B}} \\
\addlinespace[2pt]
Moving Balloon & 1~pixel/s & \textbf{100} & \textbf{100} & \textbf{100} \\
Moving Balloon & 2~pixels/s & \textbf{100} & 0 & \textbf{100} \\
Moving Balloon & 4~pixels/s & 0 & 0 & \textbf{100} \\
Moving Balloon & One 96-pixel jump at 2~s & 0 & 99 & \textbf{100} \\
Expiring Authenticator Code & Code replaced once at 2~s & 0 & 98 & \textbf{100} \\
Displaced Button & Banner moves button at 2~s & 0 & --- & \textbf{100} \\
Moving Balloon (Select) & \makecell[l]{Named target moves among\\other balloons at 8~pixels/s} & 0 & 0 & \textbf{99} \\
\midrule
\multicolumn{5}{@{}l}{\textit{UI-TARS-1.5-7B}} \\
\addlinespace[2pt]
Moving Balloon & 8~pixels/s & 0 & --- & \textbf{69} \\
\bottomrule
\end{tabular}
\caption{\textbf{Updating a pending action can recover success when both keeping it unchanged and restarting the model fail.} The screen changes while the model chooses an action for the task shown in each row. \textbf{Keep action} executes the action chosen from the initial screenshot. \textbf{Restart model} runs the model again on a newer screenshot after detecting a change. \textbf{Update action} uses a newer screenshot to change only the click position or typed code, without another model call. Each entry counts successful trials out of 100. Success is checked when the application receives the action: the click must hit the intended target, or the entered code must match the code then displayed. Results are grouped by model. The changes marked 2~s occur 2~s after the balloon, code, or page first appears. For balloon tasks, restarting is triggered by a 6-pixel change in the target's position. A dash indicates an unreported result.}
\label{tab:revision-main}
\end{table}

\subsection{Claim 3. Our Faster Control Expands the Performance Envelope}\label{subsec:claim-3}

\begin{figure}
    \centering
    \includegraphics[width=1\linewidth]{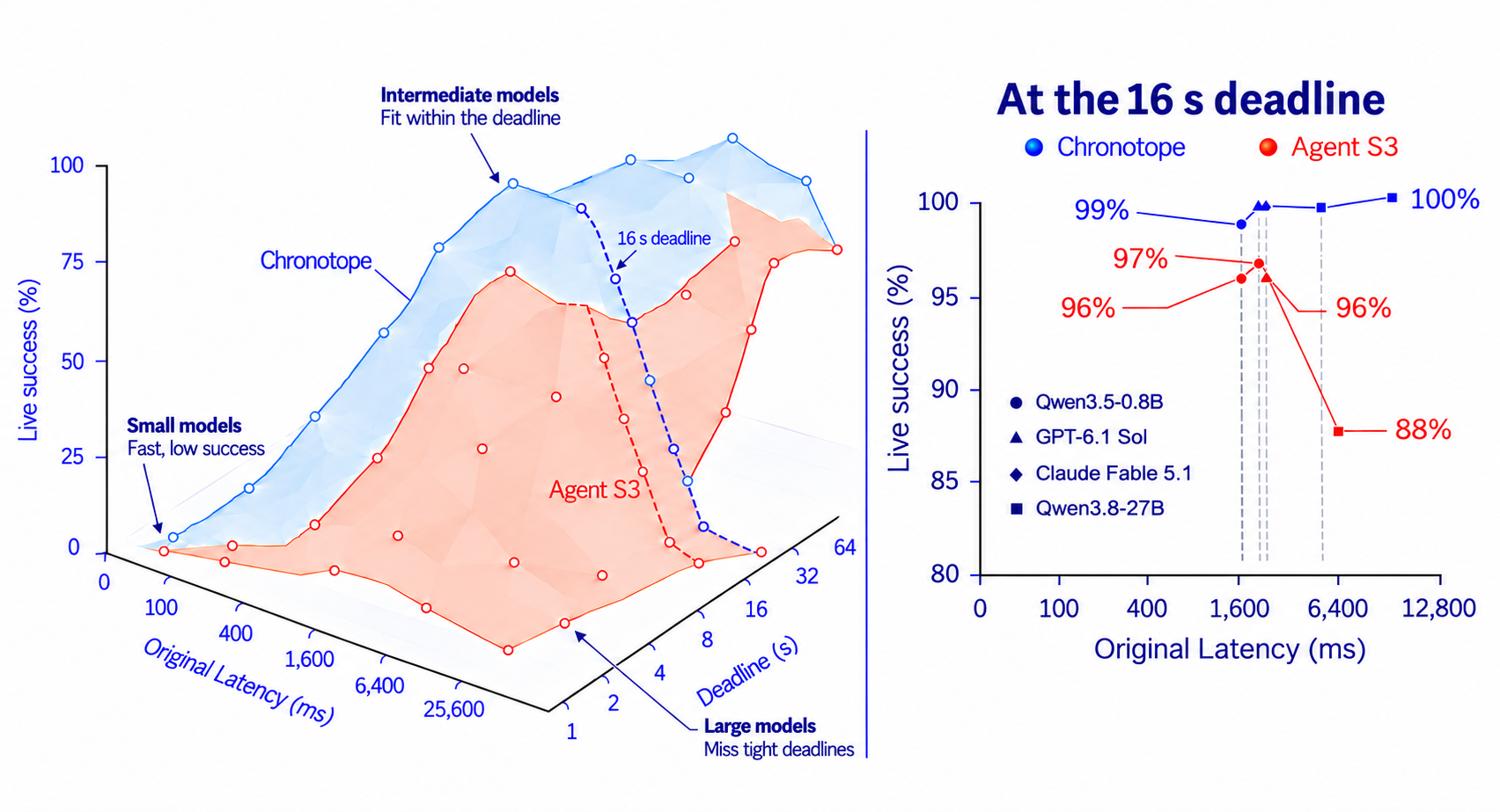}
    \caption{\textbf{Chronotope expands the performance envelope under tight response deadlines.} The main panel illustrates Live task success across original baseline end-to-end latency and response windows, with Chronotope (blue) sustaining higher success across a wider range of deadlines than Agent~S3 (red). While both controllers perform well under relaxed deadlines, their performance diverges sharply as windows tighten. Chronotope maintains 100\% success for Qwen3.8-27B at 8~s and 96\% for Qwen3.5-0.8B at 2~s, while Agent~S3 drops to zero in both cases. }\label{fig:performance-envelope}
\end{figure}

Claim 2 shows that updating a pending action can prevent failures caused by outdated action arguments, but the updated action can still fail if it reaches the application after its response deadline. Under a tight deadline, a smaller model may respond in time but choose an incorrect action, whereas a larger model may choose the correct action but respond too late, indicating a capability-latency trade-off. We thus ask whether Chronotope's faster responses let larger models achieve higher Live task success under tight deadlines, i.e., \textbf{\textit{expand their performance envelope}}. We compare Chronotope's accelerator-resident controller with the host-orchestrated Agent~S3 controller~\citep{gonzalez2026scaling}. We study Live task performance under varying deadlines for models of different sizes, and compare their original latencies under the baseline controller against Chronotope's accelerator-resident fused control. We report Live task success rate (SR), their original end-to-end latency ($E\!\rightarrow\!A$) under Agent S3, and the task response window. Figure~\ref{fig:performance-envelope} visualizes this performance envelope comparison. 


We observe that Chronotope substantially expands the Live task success attainable with each model in Figure~\ref{fig:performance-envelope}. At the same response window, Chronotope consistently envelopes Agent S3's performance envelope. At the 16~s response window, both controllers achieve high Live task success, with Chronotope reaching 100\% compared with 88\% for Agent~S3. Halving the window to 8~s reveals a sharp divergence, as Chronotope maintains perfect success while Agent~S3 drops to zero. As the window tightens further, Chronotope continues to succeed despite declining performance, until neither controller succeeds at 1~s. \textbf{\textit{Chronotope expands the performance envelope by enabling the larger model to maintain high Live task success under deadlines that the host-orchestrated controller cannot meet.}} This suggests that practitioners can favor intermediate-sized models with Chronotope to balance decision quality and responsiveness under tight deadlines. 


Chronotope achieves this expanded performance envelope by substantially reducing response latency. For the same 27B model, Chronotope achieves a 6.9$\times$ speedup in end-to-end latency over Agent~S3, bringing the response times below the 4~s deadline while Agent~S3's median remains above it. Chronotope shortens the decision path by replacing full interaction histories with compact representations and cheap action updates instead of re-grounding them through another model call, reducing token usage by up to 87.5\%. Under continuous movement, these updates consistently shorten time to commitment relative to re-grounding, with a peak reduction of 70.7\% at 8~pixels/s. This advantage persists as the target moves because Chronotope can revise an existing action without restarting inference. Chronotope maintains interaction history and performs decision reconciliation on the accelerator, reducing host-side orchestration and CPU--GPU synchronization (\S\ref{sec:fused-execution}). 

\subsection{Additional Claims}

We conducted experiments on of varying Chronotope's history capacity and on the sources of Chronotope's overall latency reduction. We claim that Chronotope's EODA/EDM mechanism provides a efficient history representation, mitigating token usage growth and latency as history capacity increases. In addition, we find that Chronotope's overall architecture obviates repeated reflection, reasoning, and grounding calls, while achieving higher success rates on DynaCU-Bench~\citep{li2026aoi}. We encourage readers to refer to Appendix~\ref{app:claim4} and \ref{app:claim5} for details.

\section{Conclusion}

GUI agents can fail despite choosing correct actions because interfaces continue changing during inference. We model the timing of environment changes, observations, decisions, and action receipt, and introduce ChronotopeGym to expose failures that response time alone cannot explain. Chronotope addresses these failures through compact interaction history, selective KEEP/UPDATE/REDECIDE meta-control, and accelerator-resident execution. On ChronotopeGym, selective meta-control matches or exceeds fixed-policy success by adapting pending actions to environmental changes. Chronotope further reduces end-to-end latency by 6.9$\times$ for a 27B model, expanding its performance envelope under tight deadlines.

\textbf{Limitations and future work.} The decision updates used in our experiments are lightweight but task-specific. Another multimodal model could infer what information in an unfinished decision should be preserved or replaced, but repeatedly invoking such a model would add inference cost and latency. Our current design therefore favors inexpensive task-specific updates while demonstrating that revising unfinished decisions can recover Live performance. Future work can study more general update mechanisms that retain this low-latency property and can additionally predict changes that occur before the action reaches the interface.
\clearpage

\bibliography{paper}
\bibliographystyle{assets/plainnat}

\newpage
\appendix
\onecolumn



\section{Extended Related Works}\label{appendix:extended_related_works}

\subsection{GUI Agents}\label{appendix:gui_agents}

GUI agents translate high-level user goals into interactions with graphical interfaces~\citep{deng2023mind2web, xie2024osworld, qin2025uitars}. A typical system receives a screenshot, sometimes together with structured interface information such as the DOM or accessibility tree, reasons about the next step, grounds that step to an interface element, and emits mouse, keyboard, touch, or hybrid GUI and API actions~\citep{zheng2024seeact, qin2025uitars, zhang2025guisurvey}. This interface-level abstraction allows generalist agents to operate across websites, desktop applications, and operating systems rather than requiring a task-specific API for every application~\citep{deng2023mind2web, xie2024osworld}.

Existing systems realize this loop in several ways. SeeAct couples multimodal planning with interface grounding~\citep{zheng2024seeact}, while Agent~S combines experience-augmented hierarchical planning with an agent--computer interface~\citep{agashe2025agents}. A complementary line trains GUI-specialized visual action models. CogAgent and UGround emphasize visual GUI understanding and grounding~\citep{hong2024cogagent,gou2024uground}, while OS-ATLAS, ShowUI, Aguvis, UI-TARS, and OpenCUA extend toward cross-platform or end-to-end action prediction~\citep{wu2025osatlas,lin2025showui,xu2025aguvis,qin2025uitars,wang2025opencua}. Despite these advances, most GUI agents still form each action from observations collected before or at the start of the current reasoning step. They improve what the agent perceives, plans, or predicts, but generally do not update an unfinished decision when relevant interface state changes during computation. As a result, an action can be correct for the state that originally supported it yet become outdated before commitment or execution. \textbf{Chronotope} addresses this limitation by keeping an unfinished decision current as new evidence arrives, preserving information that remains valid while updating information that has become outdated before commitment.

\subsection{GUI-Agent Benchmarks and Time}\label{appendix:review_bench_time}

GUI-agent benchmarks cover different parts of this stack. MiniWoB++ provides controlled browser tasks~\citep{liu2018reinforcement}. Mind2Web and Android in the Wild collect multi-step web or mobile trajectories~\citep{deng2023mind2web,rawles2023aitw}. ScreenSpot and ScreenSpot-Pro isolate GUI grounding~\citep{cheng2024seeclick,li2025screenspotpro}. OmniACT, VideoGUI, and GUI-World evaluate richer action or video-based understanding~\citep{kapoor2024omniact,lin2024videogui,chen2025guiworld}. Interactive environments then test task completion in running software, including WebArena and VisualWebArena for the web~\citep{zhou2024webarena,koh2024visualwebarena}, WorkArena for enterprise browser tasks~\citep{drouin2024workarena}, AndroidWorld for mobile applications~\citep{rawles2025androidworld}, and OSWorld, Windows Agent Arena, AgentStudio, and AssistGUI for desktop or cross-application control~\citep{xie2024osworld,bonatti2024waa,zheng2024agentstudio,gao2024assistgui}. These benchmarks have been central to progress in grounding, planning, and end-to-end task success, but their use of a live application does not by itself specify what happens to task time while the agent is computing.

Time therefore needs to be investigated as a distinct dimension of GUI-agent evaluation. Many interactive harnesses remain step-synchronous: they capture an observation, block while the agent returns an action, execute that action, and only then request the next observation. Mind2Web-Live evaluates agents against changing online websites~\citep{pan2024webcanvas}, but does not expose within-decision observation updates or score action arrival time. MiniWoB++ is an earlier exception in which its default positive reward is scaled by remaining wall-clock time, although most prior publications use a binary reward that removes this time penalty~\citep{liu2018reinforcement}. Recent benchmarks make temporal change more explicit. DynamicGUIBench scripts transient events~\citep{liu2026dynamicgui}, LivingScreen uses continuously playing interfaces~\citep{yao2026livingscreen}, DynaCU-Bench includes timed dynamic tasks~\citep{li2026aoi}, and AAPT evaluates deadline-sensitive reactions~\citep{dong2026aapt}. Table~\ref{tab:review} compares how these benchmarks represent, measure, and score time. Despite these advances, existing benchmarks either constrain interaction to a sequential observation--action loop or evaluate timing through aggregate latency and task success. As a result, they can show that timing matters without distinguishing whether correctness was lost because a change was missed, a decision became outdated, or an action arrived too late. \textbf{ChronotopeGym} addresses this gap by controlling and measuring environment change, observation availability, decision commitment, and action receipt separately, allowing timing failures to be localized within a response.

\begin{table}[htbp]
\centering
\small
\renewcommand{\arraystretch}{1.10}
\setlength{\tabcolsep}{2.5pt}
\begin{adjustbox}{max width=\linewidth,max height=0.60\textheight}
\resizebox{\linewidth}{!}{%
\begin{tabular}{@{}lccccc@{}}
\toprule
\textbf{Benchmark} &
\shortstack{\textbf{Task state}\\\textbf{during compute}} &
\shortstack{\textbf{Observation updates}\\\textbf{during decision}} &
\shortstack{\textbf{Timing}\\\textbf{recorded}} &
\shortstack{\textbf{Late action}\\\textbf{affects success?}} &
\shortstack{\textbf{Distinguishes timing}\\\textbf{failure modes?}} \\
\midrule

\multicolumn{6}{@{}l}{\textbf{Static benchmarks}} \\
\midrule

\bench{Mind2Web}{deng2023mind2web}
& Frozen
& No
& None
& No
& No \\
\cmidrule(lr){1-6}

\bench{GUI-World}{chen2025guiworld}
& Frozen
& No
& None
& No
& No \\
\cmidrule(lr){1-6}

\bench{ScreenSpot}{cheng2024seeclick}
& Frozen
& No
& None
& No
& No \\
\cmidrule(lr){1-6}

\bench{ScreenSpot-Pro}{li2025screenspotpro}
& Frozen
& No
& None
& No
& No \\
\cmidrule(lr){1-6}

\bench{OmniACT}{kapoor2024omniact}
& Frozen
& No
& None
& No
& No \\
\cmidrule(lr){1-6}

\bench{VideoGUI}{lin2024videogui}
& Frozen
& No
& None
& No
& No \\
\cmidrule(lr){1-6}

\bench{Android in the Wild}{rawles2023aitw}
& Frozen
& No
& None
& No
& No \\

\midrule
\multicolumn{6}{@{}l}{\textbf{Live sequential benchmarks}} \\
\midrule

\bench{OSWorld}{xie2024osworld}
& Advances
& No
& \shortstack{Runtime\\only}
& No
& No \\
\cmidrule(lr){1-6}

\bench{WebArena}{zhou2024webarena}
& Advances
& No
& None
& No
& No \\
\cmidrule(lr){1-6}

\bench{VisualWebArena}{koh2024visualwebarena}
& Advances
& No
& None
& No
& No \\
\cmidrule(lr){1-6}

\bench{AndroidWorld}{rawles2025androidworld}
& Advances
& No
& \shortstack{Runtime\\only}
& No
& No \\
\cmidrule(lr){1-6}

\bench{Mind2Web-Live}{pan2024webcanvas}
& Advances
& No
& None
& No
& No \\
\cmidrule(lr){1-6}

\bench{Windows Agent Arena}{bonatti2024waa}
& Advances
& No
& \shortstack{Action\\timestamps}
& No
& No \\
\cmidrule(lr){1-6}

\bench{WebVoyager}{he2024webvoyager}
& Advances
& No
& None
& No
& No \\
\cmidrule(lr){1-6}

\bench{AgentStudio}{zheng2024agentstudio}
& Advances
& No
& \shortstack{Action\\timestamps}
& \shortstack{Yes\\(optional timeout)}
& No \\
\cmidrule(lr){1-6}

\bench{Online-Mind2Web}{xue2025illusion}
& Advances
& No
& None
& No
& No \\

\midrule
\multicolumn{6}{@{}l}{\textbf{Benchmarks with explicit timing}} \\
\midrule

\bench{MiniWoB++}{liu2018reinforcement}
& \shortstack{Advances\\(timer)}
& No
& \shortstack{Remaining\\task time}
& \shortstack{Yes\\(reward)}
& No \\
\cmidrule(lr){1-6}

\bench{DynamicGUIBench}{liu2026dynamicgui}
& \shortstack{Advances\\(scripted)}
& No
& None
& No
& No \\
\cmidrule(lr){1-6}

\bench{LivingScreen}{yao2026livingscreen}
& \shortstack{Advances\\(continuous)}
& \shortstack{Yes\\(on request)}
& None
& No
& No \\
\cmidrule(lr){1-6}

\bench{DynaCU-Bench}{li2026aoi}
& \shortstack{Advances\\(timers)}
& No
& \shortstack{Decision\\latency}
& Yes
& No \\
\cmidrule(lr){1-6}

\bench{AAPT}{dong2026aapt}
& \shortstack{Advances\\(deadlines)}
& \shortstack{Yes\\(30 FPS)}
& \shortstack{Action\\receipt}
& Yes
& No \\

\midrule

\textbf{ChronotopeGym} (Ours)
& \shortstack{\textbf{Advances}\\\textbf{(controlled)}}
& \shortstack{\textbf{Yes}\\\textbf{(controlled)}}
& \textbf{\(E,O,D,A\)}
& \textbf{Yes}
& \textbf{Yes} \\

\bottomrule
\end{tabular}%
}
\end{adjustbox}

\caption{\textbf{How GUI-agent benchmarks handle time while an agent computes.}
Mind2Web, GUI-World, ScreenSpot, ScreenSpot-Pro, OmniACT, VideoGUI, and Android in the Wild use static data, so task state cannot change during an unfinished response. \textbf{ChronotopeGym} instead allows environment state to change while the agent computes.
OSWorld, WebArena, VisualWebArena, AndroidWorld, Mind2Web-Live, Windows Agent Arena, WebVoyager, AgentStudio, and Online-Mind2Web evaluate GUI interaction as a \textbf{sequential, synchronous} \(O \rightarrow A \rightarrow O \rightarrow A \rightarrow \cdots\) loop, where each observation is followed by one action before the next observation is obtained. \textbf{ChronotopeGym} instead allows observation and action events to occur independently in time, exposing timing failures that this strict alternation cannot represent.
MiniWoB++ makes remaining task time affect its default reward, but does not identify where within a response correctness is lost. \textbf{ChronotopeGym} separately measures environment change, observation availability, decision commitment, and action receipt.
DynamicGUIBench and LivingScreen introduce explicit temporal changes, but do not evaluate whether an action arrives before the relevant state changes. In contrast, \textbf{ChronotopeGym} measures action receipt against the evolving task state.
DynaCU-Bench and AAPT make latency consequential, so they can show that acting later changes overall success. However, overall success conflates whether failure arose because the agent missed a change, committed an outdated decision, or delivered an action too late. \textbf{ChronotopeGym} separates these cases by locating failure relative to environment change, observation, decision commitment, and action receipt.
}\label{tab:review}
\end{table}


\subsection{Latency and Temporal Control in GUI Agents}\label{appendix:latency_control}

Recent methods address temporal failures at different points in the interaction path. We organize them by their \textbf{temporal intervention point}: \textbf{observation-path methods}, which change when or how observations are acquired and processed,  \textbf{anticipatory methods}, which predict future interface states before execution, and \textbf{decision-path acceleration methods}, which reduce or move computation away from the online response path.

\textbf{Observation-path methods} improve what the agent can observe or how efficiently observations are processed. \citet{li2026aoi} decouple observation from action so transient visual and audio evidence can be captured between model calls, while \citet{lin2025showui} reduces visual-token computation through UI-guided token selection. However, increasing observation frequency or reducing perception cost does not determine whether newly available evidence can revise a decision already in progress. Consequently, an agent may observe a relevant change yet still execute a decision formed from an earlier state. Chronotope addresses this limitation by representing observation availability and decision commitment as separate events, allowing us to test whether newly observed evidence reaches and revises an unfinished decision.

\textbf{Anticipatory methods} compensate for future interface change by predicting what may happen before execution. \citet{cao2026mobiledreamer} predicts post-action interface states with a learned world model, while \citet{liang2026anticipatory} trains agents to forecast short-horizon trajectories before execution. However, prediction concerns expected future states rather than the actual changes that occur while inference and execution are underway. Consequently, prediction error or unmodeled change can still invalidate a planned action without revealing when correctness was lost. Chronotope addresses this limitation by representing environment change independently from observation and decision events, allowing correctness to be evaluated against the state that actually evolves during execution.

\textbf{Decision-path acceleration methods} reduce the amount of computation required after relevant state becomes available. \citet{zhang2025ufo2} amortizes model calls through speculative multi-action planning, \citet{zhong2026actionengine} reuses an explored state-machine representation to synthesize executable programs, and \citet{dong2026aapt} precompute bounded conditional policies so expensive decoding need not occur after a transient event appears. However, reducing decision latency changes how quickly an action can be produced without identifying which temporal boundary caused a remaining failure. Consequently, a failed action may still result from missing evidence, an outdated decision, invalidation after commitment, or arrival after the response opportunity has closed. Chronotope addresses this limitation by separating environment change, observation availability, decision commitment, and action receipt, so failures can be attributed to the event boundary at which correctness is lost.


\section{ChronotopeGym Tasks and Experimental Design}\label{appendix:task-protocol}

\begin{figure}[p]
\centering
\setlength{\tabcolsep}{0pt}
\renewcommand{\arraystretch}{1.0}
\begin{adjustbox}{max width=\linewidth,max height=0.70\textheight}
\begin{tabular}{@{}>{\raggedright\arraybackslash}m{0.23\linewidth}@{\hspace{0.02\linewidth}}m{0.75\linewidth}@{}}
\textbf{Disappearing Notification}\par\vspace{3pt}
{\footnotesize Opportunity closes}
&
\includegraphics[width=\linewidth]{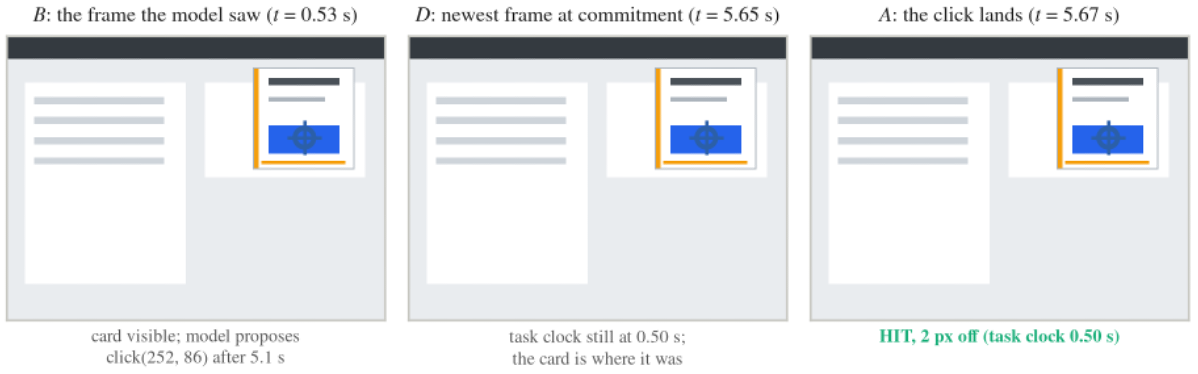}
\\[10pt]
\textbf{Moving Balloon}\par\vspace{3pt}
{\footnotesize Target position changes}
&
\includegraphics[width=\linewidth]{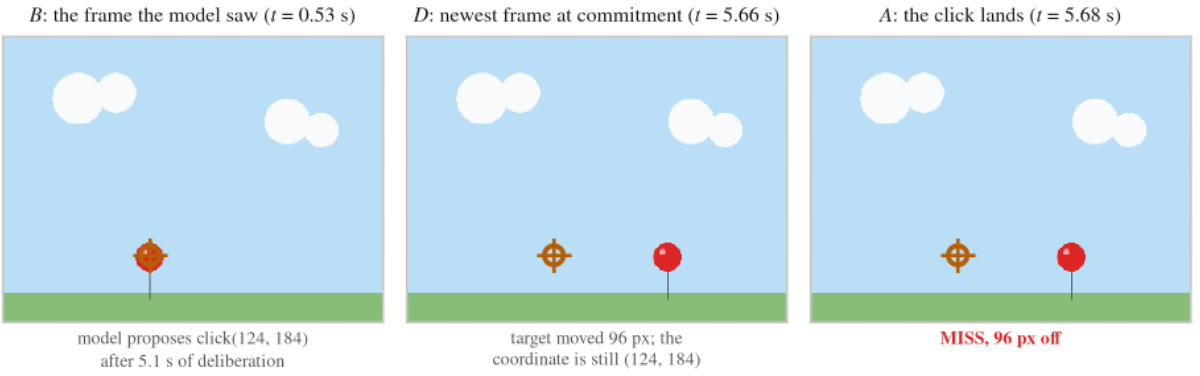}
\\[10pt]
\textbf{Expiring Authenticator Code}\par\vspace{3pt}
{\footnotesize Required value changes}
&
\includegraphics[width=\linewidth]{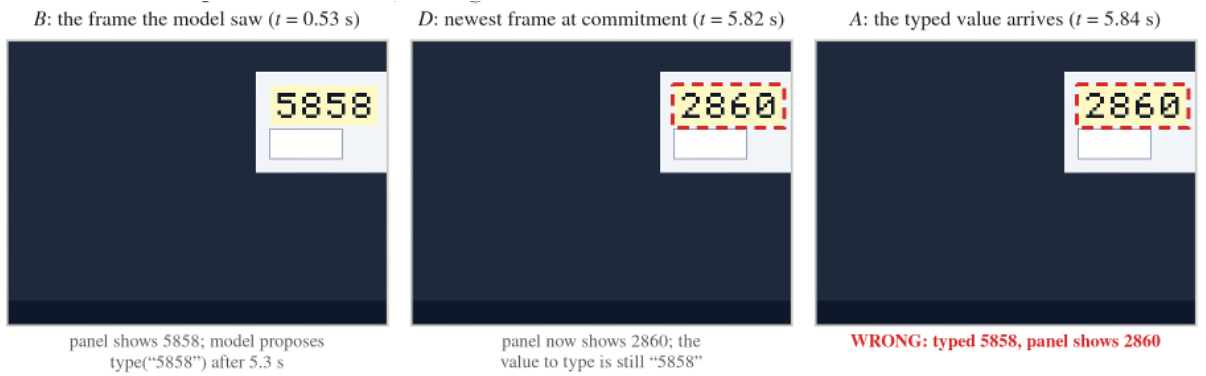}
\\[10pt]
\textbf{Displaced Button}\par\vspace{3pt}
{\footnotesize Page layout changes}
&
\includegraphics[width=\linewidth]{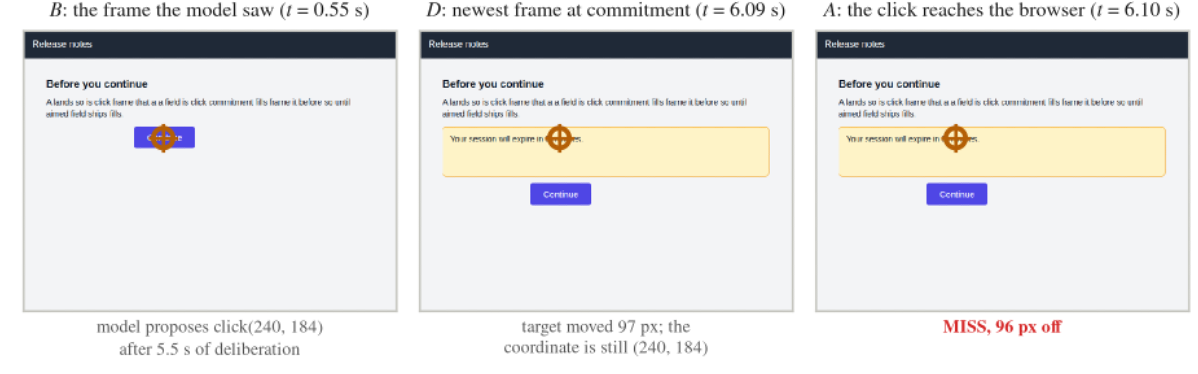}
\\[10pt]
\textbf{Reversing Trading Signal}\par\vspace{3pt}
{\footnotesize Required action changes}
&
\includegraphics[width=\linewidth]{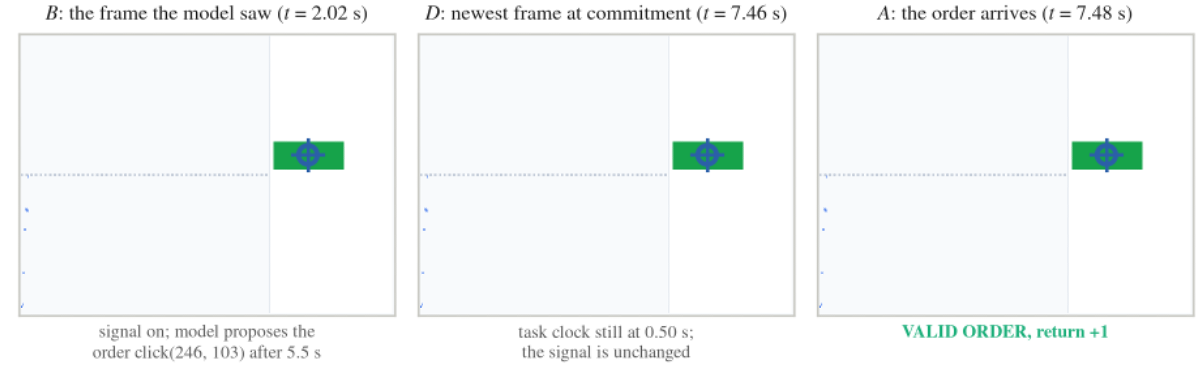}
\end{tabular}
\end{adjustbox}
\caption{\textbf{Five ChronotopeGym tasks illustrating how interface changes can invalidate an agent's response.}
Each row shows the interface initially observed by the agent (left), its changed state during deliberation (middle), and the outcome when the action reaches the application (right).
The tasks investigate closing response opportunities, moving targets, changing input values, layout-induced displacement, and changes to the required action.}\label{fig:gym}
\end{figure}

To study how an agent can choose the right action yet fail because the interface changes, we designed the five ChronotopeGym tasks introduced in \S\ref{sec:temporal-conditions}. The tasks distinguish four reasons for which an initially correct action may fail, including arriving after the opportunity closes, aiming at an outdated location, entering an outdated value, or no longer being the action the task requires. These distinctions matter because a pending action may need to be delivered sooner, updated with new information, or replaced altogether. We vary when and how the interface changes within each task to measure these failure mechanisms separately. Figure~\ref{fig:gym} illustrates the five tasks through representative GUI interactions.


\subsection{Task Design and Variants}

\paragraph{Disappearing Notification.}

The agent must click a temporary button before it disappears. A click can be correct for the observed screenshot but fail if the button disappears during inference or action delivery. A click succeeds only if the application receives it inside the button before the response window closes. The button's position and function remain unchanged while it is available, allowing us to study a closing response window without also changing the action the agent must take.

In the Card variant, a notification card with a blue button disappears when its response window closes. In the Video Player variant, a video is already playing when a sponsor banner and a temporary skip button appear. Unlike the Card variant, the skip button can appear while the agent is still completing an earlier decision, leaving less time to respond before the button disappears. In both variants, we test response windows of 1, 2, 4, 8, and 16~s to determine when the agent's response time exceeds the time available to act. For example, with a 4~s response window, a click produced after 5.1~s succeeds in Paused because the task clock stops during deliberation, but fails in Live because the notification has already disappeared.

\paragraph{Moving Balloon.}
The agent must click a balloon that can move while the agent thinks. The agent can aim at the balloon's position in the screenshot and still miss if the balloon moves before the click arrives. A click succeeds only if it lands inside the intended balloon when the application receives it. By varying when and how fast the balloon moves, we measure how its displacement during the agent's response affects click accuracy.

In the Drift variant, the balloon moves horizontally at a constant speed throughout the agent's response. We test a stationary balloon (0~pixels/s) and speeds of 1, 2, 4, 8, and 16~pixels/s to identify when the balloon moves far enough during the response for the original click to miss. In the Jump variant, the balloon shifts 96~pixels and then remains stationary, allowing us to compare updating the pending click with restarting inference. We trigger the jump at 2~s during deliberation or at 14~s after the agent's typical response, testing whether movement before versus after the click changes its outcome. In the Select variant, the agent must identify and click the balloon named in the instruction while other balloons cross its path. This distinguishes choosing the wrong balloon from choosing the right one but clicking at its previous position.

\paragraph{Expiring Authenticator Code.}

The agent must type the displayed four-digit verification code into a text field. The code may change while the agent thinks, so the value it first reads may be wrong by the time it is entered. The agent succeeds only if the entered code matches the code displayed when the application receives it. This task tests whether the agent can preserve its decision to type into the same field while updating only the code value.

In the early-replacement condition, the code changes 2~s after appearing, during deliberation, allowing us to compare retaining the original value, updating it before commitment, and restarting the decision. As controls, we either replace the code 14~s after it appears, well after the agent's typical response, or leave it unchanged throughout the response window. Comparing early replacement with these controls distinguishes failures caused by the code changing during deliberation from errors in reading or entering an otherwise valid code. As shown in Figure~\ref{fig:gym}, the code changes from 5858 to 2860 while the agent reasons, but the agent still enters the outdated value 5858.

\paragraph{Displaced Button.}

The agent must click the Continue button on a form page. A notice banner can appear above the button during deliberation, pushing it downward. A click succeeds only if it lands inside the button when the application receives it. Unlike Moving Balloon, this task tests whether changes elsewhere on the page can move the target and cause an otherwise correct click to miss.

In the early-insertion condition, the banner appears 2~s after the page is displayed while the agent is deliberating, allowing us to compare retaining the original click coordinate, updating it before commitment, and restarting inference. In the late-insertion condition, the banner appears 14~s after the page is displayed, providing a control in which the button remains at its original position throughout the agent's typical response. Comparing the two conditions distinguishes misses caused by the button moving during the response from errors in locating the button initially. Figure~\ref{fig:gym} shows a click aimed at the Continue button's original position landing on the newly inserted banner instead.

\paragraph{Reversing Trading Signal.}
The agent sees a simulated price chart and must buy, sell, or wait according to a prescribed trading rule. A buy signal can turn into a sell signal while the agent is deciding or while its order is being delivered. The task tests whether the agent follows the current trading signal, not whether it predicts future prices. An action succeeds only if it matches the signal when it reaches the application. Unlike a moving target or changing code, a signal reversal can make the agent's choice of action itself incorrect, rather than merely changing its coordinates or input values.

In the Fixed Signal condition, the required action remains the same throughout the agent's response. If the signal reverses before decision commitment \(D\), an order based on the earlier signal is already incorrect when committed. If the signal instead reverses after commitment but before action receipt \(A\), an order that was correct when committed can become invalid before reaching the application. If the reversal occurs after action receipt, the same order remains valid because it reached the application before the signal changed. In a scripted-controller experiment, we hold the committed order and its commitment and receipt times fixed, varying only when the signal reverses to isolate the effect of environment-change timing on action correctness. In a second experiment, we hold the order, commitment time, and signal reversal fixed while varying the action-delivery delay from 0 to 40~ms, testing whether the same order succeeds or fails depending on when it arrives.

\subsection{Metrics}

\paragraph{Success rate.}
Success rate is the proportion of trials in which the action satisfies the task-specific success criterion at application receipt \(A\). An action that was correct for the observed screenshot but no longer satisfies that criterion when received counts as a failure. We report successful trials out of the total number evaluated, separately for each task, agent configuration, and timing condition. We use Wilson 95\% confidence intervals for success rates~\citep{brown2001interval}.

\paragraph{Task return.}
Mean task return, \(\bar G\), complements binary success by measuring the outcome in task-specific units. For the Card variant of Disappearing Notification and for Moving Balloon, return is the negative click error in pixels, measured against the target's position at receipt. If the target is absent at receipt or no click is received, the error is set to the display diagonal. For the Video Player variant of Disappearing Notification, return is the negative number of seconds of sponsor content watched. For Reversing Trading Signal, a received order earns \(+1\) if it matches the current signal and \(-1\) otherwise. A trial with no order receives a return of 0. We average returns across trials within each condition. Higher values indicate better outcomes within the same task.

\paragraph{Timing intervals.}
We report the elapsed time from environment change to observation availability \(E\!\rightarrow\!O\), from observation availability to decision commitment \(O\!\rightarrow\!D\), and from commitment to application receipt \(D\!\rightarrow\!A\). We also report the total response time \(E\!\rightarrow\!A\). These measurements distinguish delays in obtaining an observation, producing a committed action, and delivering that action to the application.

\paragraph{Receipt-time slack.}
For tasks with a response window, receipt-time slack measures how much time remains before the window closes when the action reaches the application. Positive slack indicates arrival before the deadline. Negative slack indicates arrival after it.

\paragraph{Supporting observation and revision measurements.}
We record the source timestamp of the observation supporting the committed action to distinguish the effect of newer evidence from that of a shorter response time. For Moving Balloon, we also report the change from the proposed click coordinate to the committed coordinate and the number of model calls. These measurements show how much the action was revised and whether the revision required another model invocation.

\section{Algorithm and Complexity}\label{appendix:chronotope-algorithm}

\paragraph{Overview.}
Algorithm~\ref{alg:chronotope} expresses Chronotope as an asynchronous control procedure. Observation admission continues while the planner computes a provisional action. The controller maintains the admitted interaction history and checks the provisional action against newer evidence before commitment. If the intended operation remains appropriate, the controller preserves supported arguments, revises outdated arguments, and resolves arguments left open for late binding. If the operation itself is no longer appropriate, the controller starts a new planning call. These operations act on the provisional action rather than assuming that an external planner's hidden states or generated tokens can be edited in place.

\paragraph{Interaction history and commitment.}
The history $\mathcal H$ contains observation records, decision commitments, available action-receipt reports, and their dependencies. Environment changes are inferred from observations unless independently exposed to the controller. Evaluator-only environment timestamps and task manifests are not controller inputs. An action receipt is recorded only when receipt information becomes available, not when delivery is requested. Repeated screenshot content can share image storage without deleting distinct observation timestamps or receipt records. The snapshot $\mathcal H_*$ denotes a consistent admitted history used by the updater. It does not imply access to every physical change before commitment. The task-specific resolver may update an argument or leave it unresolved when the available evidence is insufficient. Commitment fixes the resulting action at $D$. Subsequent receipt at $A$ remains a separate event, and changes after the updater's snapshot can still invalidate the action.

\begin{algorithm}[t]
\caption{Chronotope: asynchronous observation and pre-commit action reconciliation}
\label{alg:chronotope}
\begin{algorithmic}[1]
\Require Task $g$, planner $\pi_\theta$, task-specific checks and argument resolvers
\State Initialize event dependency matrix $\mathcal H \gets \varnothing$
\State Provisional action $\tilde a \gets \bot$, planner handle $f \gets \bot$
\While{\Call{TaskActive}{$g$}}
    \State $\mathcal X \gets \Call{AvailableInputs}{}$
    \State $\mathcal H \gets \Call{UpdateHistory}{\mathcal H,\mathcal X}$
    \If{$f=\bot$ and $\tilde a=\bot$ and \Call{DecisionNeeded}{$g,\mathcal H$}}
        \State $f \gets \Call{StartPlanner}{\pi_\theta,g,\mathcal H}$ \Comment{Nonblocking}
    \EndIf
    \If{$f\neq\bot$ and \Call{PlannerReady}{$f$}}
        \State $\tilde a \gets \Call{CollectProposal}{f}$, $f \gets \bot$
    \EndIf
    \If{$\tilde a\neq\bot$}
        \State $\mathcal H_* \gets \Call{Snapshot}{\mathcal H}$
        \If{\Call{OperationInvalid}{$\tilde a,\mathcal H_*$}}
            \State $\tilde a \gets \bot$
            \State $f \gets \Call{StartPlanner}{\pi_\theta,g,\mathcal H_*}$ \Comment{Re-decision}
        \Else
            \State $J \gets \Call{StaleOrUnboundArguments}{\tilde a,\mathcal H_*}$
            \For{each argument $j\in J$}
                \State $\tilde a_j \gets \Call{Resolve}{\tilde a,j,\mathcal H_*}$ \Comment{Revise or late bind}
            \EndFor
            \If{\Call{CompleteAndSupported}{$\tilde a,\mathcal H_*$}}
                \State $(a,d) \gets \Call{Commit}{\tilde a,\mathcal H_*}$ \Comment{$d$ records $D$ and evidence used}
                \State $\mathcal H \gets \Call{AppendDecision}{\mathcal H,d}$
                \State \Call{EnqueueDelivery}{$a$} \Comment{Does not imply receipt at $A$}
                \State $\tilde a \gets \bot$
            \EndIf
        \EndIf
    \EndIf
    \State \Call{AwaitNextEvent}{} \Comment{Input, planner completion, or task event}
\EndWhile
\end{algorithmic}
\end{algorithm}

\paragraph{Time complexity.}
Let $h$ denote the maximum number of retained fixed-size event records, $F$ the decoded frame size, and $K$ the provisional-action size. With preallocated storage and precomputed metadata, incremental dependency maintenance costs $\mathcal{O}(h)$ per inserted event. A full scan of the dependency matrix and action bookkeeping cost $\mathcal{O}(h^2+K)$ per reconciliation. Bytewise duplicate detection costs $\mathcal{O}(F)$ per received frame. Perception, semantic checks, argument resolution, and planner inference contribute additional costs. Re-decision adds another planner invocation.

\paragraph{Space complexity.}
With at most $B$ retained image buffers, additional controller memory is $\mathcal{O}(BF+h^2+K+S_{\mathrm{scratch}})$, where $S_{\mathrm{scratch}}$ includes feature buffers and temporary perception and resolver storage. Model parameters, activations, and inference caches contribute additional memory. Bounding controller memory independently of session duration requires limits on retained data and temporary state, including data referenced by active snapshots.

\paragraph{Host--device communication.}
For $J$ intermediate host--device transfers carrying at most $S_c$ bytes each, intermediate communication is $\mathcal{O}(JS_c)$. A fully resident control path eliminates these transfers. Observation input, finalized-action output, receipt information, and any runtime copies or managed-memory migrations must be counted separately. The latency saving depends on the transferred buffers and per-transfer overhead.

\paragraph{Device-memory traffic.}
With a bounded number of metadata passes, each reconciliation accesses $\mathcal{O}(h^2+K)$ logical bytes, excluding visual processing and model inference. Actual DRAM traffic depends on cache behavior. Accelerator residency does not imply that intermediate state remains on chip. Kernel fusion can avoid intermediate DRAM writes and reads when that state remains on chip without spilling. These work and traffic bounds do not directly predict elapsed latency because computation and transfers can overlap, while scheduling and synchronization add overhead.

\section{Additional Results}\label{appendix:extra-results}

\subsection{Claim 4: EODA Enables Efficient Interaction History}\label{app:claim4}

EODA (\S\ref{sec:interaction-state}) concisely represents interaction history by organizing dependencies among environment events, observations, decisions, and action receipts. Unlike conventional chronological text history, EODA explicitly represents relationships between previous interactions and their outcomes, allowing the planner to use past actions when generating subsequent decisions. The representation maintains a bounded set of changed observations with GPU-resident textual and visual embeddings, allowing relevant observations to be reused without recomputation. To examine the effect of retained history, we vary capacity $N\in\{0,1,2,4,8,16\}$ on DynaCU-Bench Standard and report task success, input token consumption, and median end-to-end latency per sent action.

\begin{table}[t]
\centering
\small
\setlength{\tabcolsep}{4pt}
\renewcommand{\arraystretch}{1.12}
\begin{adjustbox}{max width=\linewidth}
\begin{tabular}{@{}rrrr@{}}
\toprule
\shortstack{Capacity\\$N$} &
\shortstack{DynaCU-Bench\\Standard SR} &
\shortstack{Input tokens\\per episode} &
\shortstack{End-to-end latency\\per sent action, median (ms)} \\
\midrule
0  & 41/100 (41\%) & 54,996 & 1,574 \\
1  & 45/100 (45\%) & 56,586 & 1,596 \\
2  & 48/100 (48\%) & 58,136 & 1,603 \\
4  & \textbf{49/100 (49\%)} & 63,267 & 1,616 \\
8  & 46/100 (46\%) & 71,227 & 1,668 \\
16 & 47/100 (47\%) & 84,435 & 1,732 \\
\bottomrule
\end{tabular}
\end{adjustbox}
\caption{\textbf{Effect of interaction-history capacity on DynaCU-Bench Standard.} We vary the retained history capacity and report task success, input tokens per episode, and median end-to-end latency per sent action.}
\label{tab:eoda-history-capacity}
\end{table}

As shown in Table~\ref{tab:eoda-history-capacity}, increasing history capacity from $N=0$ to $N=4$ improves task success by 19.5\%, while increasing input token consumption by 15.0\% and median per-action latency by only 2.7\%. Beyond $N=4$, larger history capacities do not yield further improvements in task success, despite continued increases in token consumption and latency. At $N=16$, token consumption increases by 33.5\% relative to $N=4$, while task success slightly decreases. \textbf{\textit{These results suggest that a modest history capacity provides a favorable balance between retained context, task success, and computational efficiency.}}

\subsection{Claim 5: Chronotope Reduces End-to-End Latency through Architectural Efficiency}\label{app:claim5}

Chronotope's main latency advantage stems from changing how control is performed. Traditional reflection-based agents such as Agent S3 repeatedly use model inference for verification, screenshot analysis, planning, and target localization. In contrast, Chronotope maintains interaction history, decision dependencies, and pending actions through EODA/EDM and selective meta-control (KEEP, UPDATE, REDECIDE). This allows it to reuse previous decisions, reconcile pending actions with new observations, and avoid repeatedly invoking the VLM for reflection and separate grounding. As a result, Chronotope substantially reduces expensive model-side computation while preserving the ability to adapt actions online.

\begin{figure}[ht]
\centering
\includegraphics[width=0.58\linewidth]{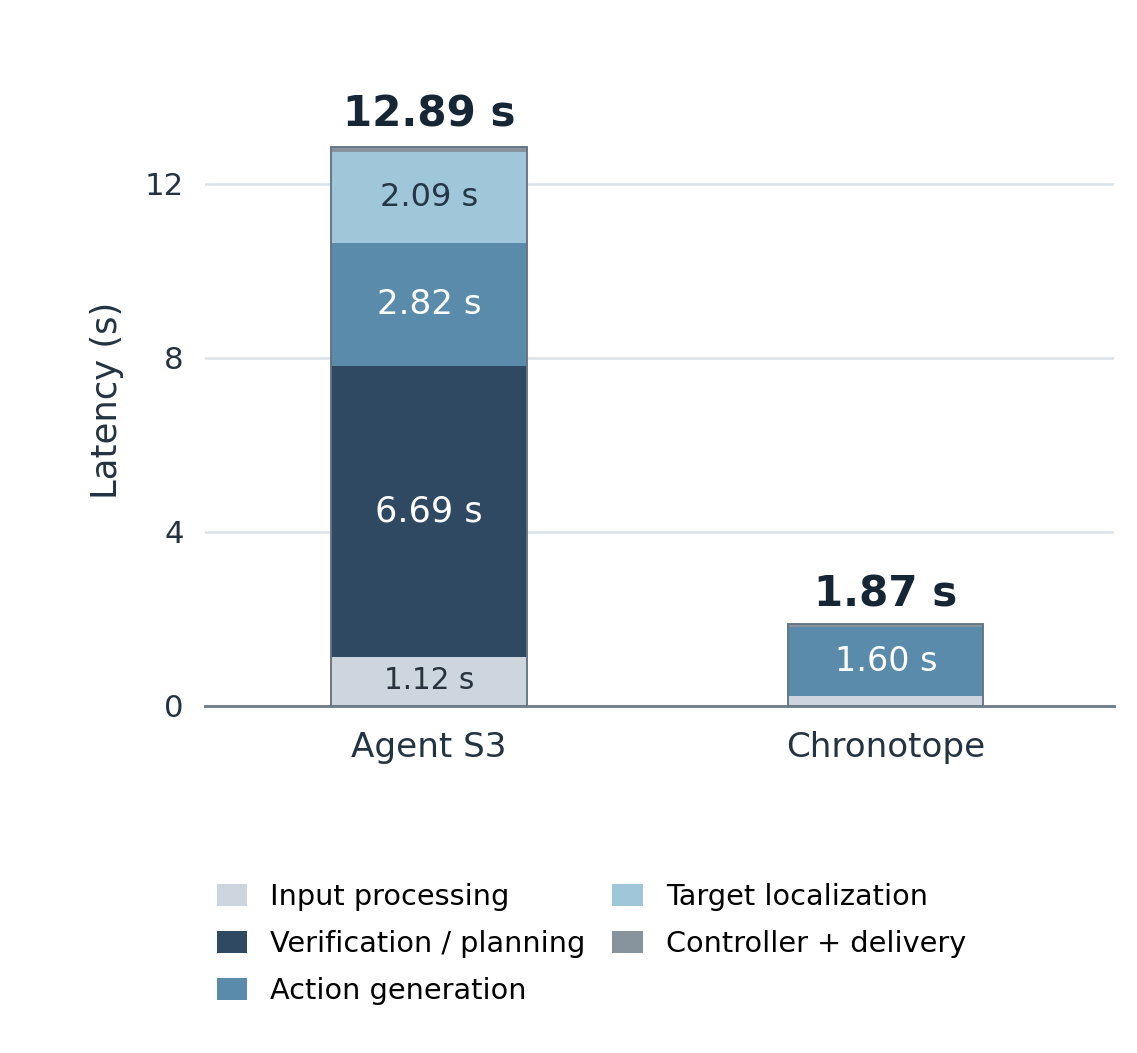}
\caption{Functional latency breakdown of Agent S3 and Chronotope across 100 paired Live trials using Qwen3.8-27B. The dominant overhead in Agent S3 comes from repeated model-based reflection, analysis, and grounding, which Chronotope largely avoids through compact interaction state, selective meta-control, and lightweight target localization.}
\label{fig:architectural_latency}
\end{figure}

We profile both GUI agents across 100 paired Live trials using Qwen3.8-27B. Figure~\ref{fig:architectural_latency} compares their functional latency breakdowns. Agent S3 spends \textbf{51.9\%} of its end-to-end latency on verification, analysis, and planning prose, \textbf{16.2\%} on target localization through two separate UI-TARS grounding calls, and \textbf{8.7\%} on input processing. By contrast, Chronotope uses compact inputs, lightweight GPU-based target localization, and avoids the repeated reflection and grounding stages that dominate Agent S3's latency. Overall, Chronotope reduces end-to-end latency by \textbf{85.5\%}.

This functional decomposition helps explain the source of the full-system speedup. Of the total \textbf{85.5\%} reduction, shorter model generation contributes \textbf{61.4 percentage points}, faster target localization contributes \textbf{16.2 points}, and shorter input processing contributes \textbf{7.1 points}. Notably, Agent S3 spends approximately \textbf{74\%} of its end-to-end latency on textual verification, screenshot analysis, planning, and action-description generation. Chronotope removes most of this overhead by representing interaction state directly in its controller and generating only compact executable actions.

Importantly, this improvement does not primarily reflect faster model serving. Agent S3 uses vLLM with longer prompts and separate grounding calls, while Chronotope uses in-process TensorRT-Edge-LLM with compact prompts and lightweight grounding. Chronotope's token generation is in fact slightly slower, at 84.4 versus 80.1 ms/token. The measured speedup therefore comes chiefly from reducing unnecessary model-side work, rather than from higher raw generation throughput.

The same architectural distinction is relevant in multi-step settings. Across 16 multi-step DynaCU-Bench episodes, Agent S3 spends an average of 15,345 ms per turn on generation, 6,938 ms on reflection, 2,509 ms on context processing, and 1,777 ms on grounding. Generation and reflection together account for \textbf{83.6\%} of its model-call time. These measurements characterize the recurring overhead that Chronotope is designed to avoid by maintaining compact interaction state and selectively updating pending actions instead of repeatedly reconstructing context through VLM inference.

Accelerator residency provides a further but smaller improvement. In a matched CPU-versus-GPU Chronotope ablation that holds the planner, EODA/EDM representation, meta-control policy, updater, and execution conditions fixed, GPU residency reduces controller-path latency from 67.2 ms to 19.9 ms, a \textbf{70.4\% reduction}. This corresponds to an additional \textbf{2.4\% reduction in end-to-end latency} for Qwen3.8-27B. Thus, accelerator placement helps, but it explains only a small fraction of the much larger full-system speedup over Agent S3.

Overall, Chronotope reduces end-to-end latency primarily through its overall architecture, including compact interaction representation, selective decision reconciliation, compact action generation, and lightweight target localization together avoid the repeated reflection and grounding calls required by traditional reflection-based GUI agents.

\end{document}